\documentclass[sigconf]{aamas}  %
\usepackage{luatex85}
\usepackage[textsize=tiny]{todonotes}
\usepackage{amsmath}
\usepackage{amssymb}
\usepackage{mathtools}
\usepackage{enumitem}
\usepackage{bbm}
\usepackage[skip=3pt]{caption}
\usepackage{subcaption}
\usepackage{adjustbox}
\usepackage{algorithm}
\usepackage[noend]{algorithmic}
\usepackage{footnote}
\usepackage{tablefootnote}
\usepackage{multirow}
\usepackage{booktabs}
\usepackage[tight-spacing=true]{siunitx}
\usepackage{tikz,tkz-graph}
\setcopyright{ifaamas}  %
\acmDOI{}  %
\acmISBN{}  %
\acmConference[AAMAS'19]{Proc.\@ of the 18th International Conference on Autonomous Agents and Multiagent Systems (AAMAS 2019)}{May 13--17, 2019}{Montreal, Canada}{N.~Agmon, M.~E.~Taylor, E.~Elkind, M.~Veloso (eds.)}  %
\acmYear{2019}  %
\copyrightyear{2019}  %
\acmPrice{}  %

\usetikzlibrary{
  graphs,
  graphs.standard,
  positioning,chains,fit,shapes,calc
}
\newcommand\modprim{\ensuremath{\hat{\mathcal{T}}}} %

\definecolor{nice-red}{HTML}{E41A1C}
\definecolor{nice-orange}{HTML}{FF7F00}
\definecolor{nice-yellow}{HTML}{FFC020}
\definecolor{nice-green}{HTML}{4DAF4A}
\definecolor{nice-blue}{HTML}{377EB8}
\definecolor{nice-nice-red}{HTML}{984EA3}

\begin{document}
\title{Model Primitive Hierarchical Lifelong Reinforcement Learning }

\author{Bohan Wu}  %
\affiliation{%
\institution{Columbia University}
}
\email{bw2505@columbia.edu}
\author{Jayesh K. Gupta}
\affiliation{%
\institution{Stanford University}
}
\email{jkg@cs.stanford.edu}
\author{Mykel J. Kochenderfer}
\affiliation{%
\institution{Stanford University}
}
\email{mykel@stanford.edu}

\begin{abstract}
  Learning interpretable and transferable subpolicies and performing task decomposition from a single, complex task is difficult. 
  Some traditional hierarchical reinforcement learning techniques enforce this decomposition in a top-down manner, while meta-learning techniques require a task distribution at hand to learn such decompositions. 
  This paper presents a framework for using diverse suboptimal world models to decompose complex task solutions into simpler modular subpolicies.
  This framework performs automatic decomposition of a single source task in a bottom up manner, concurrently learning the required modular subpolicies as well as a controller to coordinate them.
  We perform a series of experiments on high dimensional continuous action control tasks to demonstrate the effectiveness of this approach at both complex single task learning and lifelong learning.
  Finally, we perform ablation studies to understand the importance and robustness of different elements in the framework and limitations to this approach.
\end{abstract}

\maketitle

\section{Introduction}
In the lifelong learning setting, we want our agent to solve a \textit{series} of related tasks drawn from some task distribution rather than a single, isolated task.
Agents must be able to \textit{transfer} knowledge gained in previous tasks to improve performance on future tasks.
This setting is different from multi-task reinforcement learning~\cite{tanaka2003multitask,wilson2007multi,teh2017distral} and various meta-reinforcement learning settings~\cite{mlsh,finn2017model}, where the agent jointly trains on multiple task environments.
Not only do such non-incremental settings make the problem of discovering common structures between tasks easier, they allow the methods to ignore the problem of catastrophic forgetting~\cite{mccloskey1989-nh}, which is the inability to solve previous tasks after learning to solve new tasks in a sequential learning setting.

Our work takes a step towards solutions for such incremental settings. 
We draw on the idea of modularity~\cite{Neumann2014-mq}.
While learning to perform a complex task, we force the agent to break its solution down into simpler subpolicies instead of learning a single monolithic policy.
This decomposition allows our agent to rapidly learn another related task by transferring these subpolicies.
We hypothesize that many complex tasks are heavily structured and hierarchical in nature.
The likelihood of transfer of an agent's solution increases if it can capture such shared structure.

A key ingredient of our proposal is the idea of world models~\cite{Keller2012,leinweber2017,Ha2018WorldModels} --- transition models that can predict future sensory data given the agent's current actions.
The world however is complex, and learning models that are consistent enough to plan with is not only hard~\cite{Talvitie2017SelfCorrectingMF}, but planning with such one-step models is suboptimal~\cite{onestepplanbad18}.
We posit that the requirement that these world models be good predictors of the world state is unnecessary, provided we have a multiplicity of such models.
We use the term \textit{model primitives} to refer to these suboptimal world models. Since each model primitive is only relatively better at predicting the next states within a certain region of the environment space, we call this area the model primitive's \textit{region of specialization}.

Model primitives allow the agent to decompose the task being performed into subtasks according to their regions of specialization and learn a specialized subpolicy for each subtask.
The same model primitives are used to learn a gating controller to select, improve, adapt, and sequence the various subpolicies to solve a given task in a manner very similar to a mixture of experts framework~\cite{moesurvey14}.

Our framework assumes that at least a subset of model primitives are useful across a range of tasks and environments.
This assumption is less restrictive than that of successor representations~\cite{Dayan1993-ip,barreto2017successor}.
Even though successor representations decouple the state transitions from the rewards (representing the task or goals), the transitions learned are policy dependent and can only transfer across tasks with the same environment dynamics.

There are alternative approaches to learning hierarchical spatio-temporal decompositions from the rewards seen while interacting with the environment.
These approaches include meta-learning algorithms like Meta-learning Shared Hierarchies (MLSH)~\cite{mlsh}, which require a multiplicity of pretrained subpolicies and joint training on related tasks.
Other approaches include the option-critic architecture~\cite{option-critic} that allows learning such decompositions in a single task environment.
However, this method requires regularization hyperparameters that are tricky to set. 
As observed by \citet{vezhnevets2017feudal}, its learning often collapses to a single subpolicy.
Moreover, we posit that capturing the shared structure across task-environments can be more useful in the context of transfer for lifelong learning than reward-based task specific structures.

To summarize our contributions:
\begin{itemize}
\item Given diverse suboptimal world models, we propose a method to leverage them for task decomposition.
\item We propose an architecture to jointly train decomposed subpolicies and a gating controller to solve a given task.
\item We demonstrate the effectiveness of this approach at both single-task and lifelong learning in complex domains with high-dimensional observations and continuous actions.
\end{itemize}

\section{Preliminaries}
We assume the standard reinforcement learning (RL) formulation: an \textit{agent} interacts with an \textit{environment} to maximize the expected reward~\cite{sutton1998reinforcement}.
The environment is modeled as a Markov decision process (MDP), which is defined by $\langle \mathcal{S}, \mathcal{A}, \mathcal{R}, \mathcal{T}, \gamma \rangle$ with a state space $\mathcal{S}$, an action space $\mathcal{A}$, a reward function $\mathcal{R}: \mathcal{S} \times \mathcal{A} \to \mathbb{R}$, a dynamics model $\mathcal{T}: \mathcal{S} \times \mathcal{A} \to \Pi(\mathcal{S})$, and a discount factor $\gamma \in [0, 1)$.
Here, $\Pi(\cdot)$ defines a probability distribution over a set.
The agent acts according to stationary stochastic policies $\pi: \mathcal{S} \to \Pi(\mathcal{A})$, which specify action choice probabilities for each state.
Each policy $\pi$ has a corresponding $Q_\pi: \mathcal{S}\times\mathcal{A} \to \mathbb{R}$ function that defines the expected discounted cumulative reward for taking an action $a$ from state $s$ and following the policy $\pi$ from that point onward.

\textbf{Lifelong Reinforcement Learning}:
In a lifelong learning setting, the agent must interact with multiple tasks and successfully solve each of them.
Adopting the framework from~\citet{Brunskill2014-gr}, in lifelong RL, the agent receives $\mathcal{S}, \mathcal{A}$, initial state distribution $\rho_0 \in \Pi(\mathcal{S})$, horizon $H$, discount factor $\gamma$, and an unknown distribution over reward-transition function pairs, $D$. 
The agent samples $(\mathcal{R}_i, \mathcal{T}_i) \sim D$ and interacts with the MDP $\langle \mathcal{S}, \mathcal{A}, \mathcal{R}_i, \mathcal{T}_i, \gamma \rangle$ for a maximum of $H$ timesteps, starting according to the initial state distribution $\rho_0$. 
After solving the given MDP or after $H$ timesteps, whichever occurs first, the agent resamples from $D$ and repeats.

The fundamental question in lifelong learning is to determine what knowledge should be captured by the agent from the tasks it has already solved so that it can improve its performance on future tasks. 
When learning with functional approximation, this translates to learning the right representation --- the one with the right inductive bias for the tasks in the distribution. 
Given the assumption that the set of related tasks for lifelong learning share a lot of structure, the ideal representation should be able to capture this shared structure.

\citet{Thrun1998} summarized various representation decomposition methods into two major categories. Modern approaches to avoiding catastrophic forgetting during transfer tend to fall into either category. The first category partitions the parameter space into task-specific parameters and general parameters~\cite{Rusu2016-pnn}. The second category learns constraints that can be superimposed when learning a new function~\cite{kirkpatrick2017overcoming}. 

A popular approach within the first category is to use what~\citet{Thrun1998} term as \textit{recursive} functional decomposition. 
This approach assumes that solution to tasks can be decomposed into a function of the form $f_i = h_i \circ g$, where $h_i$ is task-specific whereas $g$ is the same for all $f_i$. 
This scheme has been particularly effective in computer vision where early convolutional layers in deep convolutional networks trained on ImageNet~\cite{deng2009imagenet,sun2017revisiting} become a very effective $g$ for a variety of tasks.
However, this approach to decomposition often fails in DeepRL because of two main reasons. 
First, the gradients used to train such networks are noisier as a result of Monte Carlo sampling. 
Second, the i.i.d.\ assumption for training data often fails.

We instead focus on devising an effective \textit{piecewise} functional decomposition of the parameter space, as defined by \citet{Thrun1998}.
The assumption behind this decomposition is that each function $f_i$ can be represented by a collection of functions $h_1, \ldots, h_m$, where $m \ll N$, and $N$ is the number of tasks to learn.
Our hypothesis is that this decomposition is much more effective and easier to learn in RL.

\section{Model Primitive Hierarchical RL}
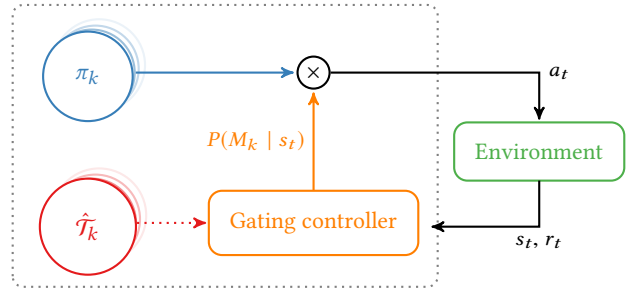
\begin{figure}[t]
  \centering
  \begin{tikzpicture}[->, >=stealth', shorten >=1pt, auto, node distance=0.8cm, thick, main/.style={draw, inner sep=8pt, outer sep=0pt, rounded corners=5pt}]
    \node[circle, fill=white, draw=nice-blue!10, inner sep=8pt] (pi_k) at (1.15, -0.85) {\color{nice-blue}$\pi_{K}$};
    \node[circle, fill=white, draw=nice-blue!30, inner sep=8pt] (pi_3) at (1.10, -0.90) {\color{nice-blue}$\pi_{3}$};
    \node[circle, fill=white, draw=nice-blue!50, inner sep=8pt] (pi_2) at (1.05, -0.95) {\color{nice-blue}$\pi_{2}$};
    \node[circle, fill=white, draw=nice-blue, inner sep=8pt] (pi_1) at (1.00, -1.00) {\color{nice-blue}  $\pi_{k}$};
    \node[circle, fill=white, draw=nice-red!10, inner sep=8pt]  (tau_k) at (1.15, -2.85) {\color{nice-red} $\mathcal{T}_K$};
    \node[circle, fill=white, draw=nice-red!30, inner sep=8pt]  (tau_3) at (1.10, -2.90) {\color{nice-red} $\mathcal{T}_3$};
    \node[circle, fill=white, draw=nice-red!50, inner sep=8pt]  (tau_2) at (1.05, -2.95) {\color{nice-red} $\mathcal{T}_2$};
    \node[circle, fill=white, draw=nice-red, inner sep=8pt]  (tau_1) at (1.00, -3.00) {\color{nice-red}  $\hat{\mathcal{T}}_k$};
    \node[circle, draw=black, inner sep=2pt] (prod) at (4.00, -0.95) {\small $\times$};
    \node[main, draw=nice-orange] (gating controller) at (4.00, -2.95) {\color{nice-orange}  Gating controller};
    \node[main, draw=nice-green] (env) at (7.00, -2.0) {\color{nice-green}  Environment};
    \node [inner sep=0pt] (hid) at (5.50, -3.00) {};
    \draw[dotted, draw=gray, rounded corners] (0, -3.8) rectangle ++(5.65, 3.75) {};
    \draw[dotted, nice-red] (tau_2) to (gating controller);
    \draw[nice-orange] (gating controller) to node [auto] {\color{nice-orange} \small $P(M_k\mid s_t)$} (prod);
    \draw[nice-blue] (pi_2) to (prod);
    \draw[] (prod) -| node [auto] {\small $a_t$} (env);
    \draw[] (env) |- node [auto] {\small $s_t, r_t$}(hid);
  \end{tikzpicture}
  \caption{Diagram of MPHRL Architecture. Solid arrows are active during both learning and execution. Dotted arrows are active only during learning.}\label{fig:framework}
\end{figure}

This section outlines the Model Primitive Hierarchical Reinforcement Learning (MPHRL) framework (Figure~\ref{fig:framework}) to address the problem of effective piecewise functional decomposition for transfer across a distribution of tasks.

\subsection{Model Primitives and Gating}
The key assumption in MPHRL is access to several diverse world models of the environment dynamics.
These models can be seen as instances of learned approximations to the true environment dynamics $\mathcal{T}$.
In reality, these dynamics can even be non-stationary.
Therefore, the task of learning a complete model of the environment dynamics might be too difficult.
Instead, it can be much easier to train multiple approximate models that specialize in different parts of the environment.
We use the term model primitives to refer to these approximate world models.

Suppose we have access to $K$ model primitives: $\modprim_k: \mathcal{S} \times \mathcal{A} \to \Pi(\mathcal{S})$.
For simplicity, we can assign a label $M_k$ to each $\modprim_k$, such that their predictions of the environment's transition probabilities can be denoted by $\modprim(s_{t+1} \mid s_t, a_t, M_k)$.

\subsubsection{Subpolicies}
The goal of the MPHRL framework is to use these suboptimal predictions from different model primitives to decompose the task space into their regions of specialization, and learn different subpolicies $\pi_k: \mathcal{S} \to \Pi(\mathcal{A})$ that can focus on these regions.
In the function approximation regime, each subpolicy $\pi_k$ belongs to a fixed class of smoothly parameterized stochastic policies $\{\pi_{\theta_k} \mid \theta_k \in \Theta\}$, where $\Theta$ is a set of valid parameter vectors.

Model primitives are suboptimal and make incorrect predictions about the next state.
Therefore we do not use them for planning or model-based learning of subpolicies directly.
Instead, model primitives give rise to useful functional decompositions and allow subpolicies to be  learned in a model-free way.

\subsubsection{Gating Controller}
Taking inspiration from the mixture-of-experts literature~\cite{moesurvey14}, where the output from multiple experts can be combined using probabilistic gating functions, MPHRL decomposes the solution for a given task into multiple ``expert'' subpolicies and a gating controller that can compose them to solve the task. 
We want this switching behavior to be probabilistic and continuous to avoid abrupt transitions.
During learning, we want this controller to help assign the reward signal to the correct blend of subpolicies to ensure effective learning as well as decomposition.

Since the gating controller's goal is to choose the subpolicy whose corresponding model primitive makes the best prediction for a given transition, using Bayes' rule we can write:
\begin{multline}
    P(M_k \mid s_t, a_t, s_{t+1}) \propto \\ {{P(M_k \mid s_t) \pi_k(a_t \mid s_t)}{\modprim(s_{t+1} \mid s_t, a_t, M_k)}}
  \label{eq:posterior}
\end{multline}
because $\pi_k(a_t \mid s_t) = \pi(a_t \mid s_t, M_k)$.

The agent only has access to the current state $s_t$ during execution.
Therefore, the agent needs to marginalize out $s_{t+1}$ and $a_t$ such that the model choice only depends on the current state $s_t$:
\begin{multline}
    P(M_k \mid s_t) = \int\limits_{s_{t+1} \in \mathcal{S}}\int\limits_{a_t \in \mathcal{A}} P(M_k \mid s_t, a_t, s_{t+1}) \\ P(s_{t+1}, a_t)da_{t} ds_{t+1}
    \label{eq:marginal}
\end{multline}

This is equivalent to:
\begin{equation} 
  P(M_k \mid s_t) = \mathbb{E}_{s_{t+1}, a_t \sim P(s_{t+1}, a_t)}\left[P(M_k \mid s_t, a_t, s_{t+1})\right]
\end{equation}

Unfortunately, computing these integrals requires expensive Monte Carlo methods. %
However, we can use an approximate method to achieve the same objective with discriminative learning~\cite{rosenbaum2015}.

We parameterize the gating controller (GC) as a categorical distribution $P_{\phi}(M_k \mid s_t) = P(M_k \mid s_t; \phi)$ and minimize the conditional cross entropy loss between $\mathbb{E}_{s_{t+1}, a_t \sim P(s_{t+1}, a_t)}\left[P(M_k \mid s_t, a_t, s_{t+1})\right]$ and $P_{\phi}(M_k \mid s_t)$ for all sampled transitions $(s_t, a_t, s_{t+1})$ in a rollout:
\begin{equation}
    \underset{\phi}{\text{minimize }} \mathcal{L}^{GC}
    \label{eq:cem}
\end{equation}
where
\begin{multline}
    \mathcal{L}^{GC} = \sum_{s_t} \sum_k -\left(\sum_{s_{t+1}} \sum_{a_t} P(M_k \mid s_{t}, a_t, s_{t+1})\right) \\ \times \log {P(M_k \mid s_t; \phi)}
    \label{eq:GC}
\end{multline}
This is equivalent to an implicit Monte Carlo integration to compute the marginal if $s_{t+1}, a_t \sim P(s_{t+1}, a_t)$.
Although we cannot query or sample from $P(s_{t+1}, a_t)$ directly, $s_t, a_t$, and $s_{t+1}$ can be sampled according to their respective distributions while we perform rollouts in the environment.
Despite the introduced bias in our estimates, we find Eq.~\ref{eq:cem} sufficient for achieving task decomposition.

\subsubsection{Subpolicy Composition}
Taking inspiration from mixture-of-experts, the gating controller composes the subpolicies into a mixture policy:
\begin{equation}
  \pi(a_t \mid s_t) = \sum_{k=1}^K P_{\phi}(M_k \mid s_t) \pi_k(a_t \mid s_t)
  \label{eq:mix}
\end{equation}

\subsubsection{Decoupling Cross Entropy from Action Distribution} 
During a rollout, the agent samples as follows:
\begin{align}
    a_t &\sim \pi(a _t \mid s_t) \\
    s_{t+1} &\sim \mathcal{T}(s_{t+1} \mid s_t, a_t)
    \label{eq:approxsamp}
\end{align}

The $\pi_k$ from Eq.~\ref{eq:posterior} gets coupled with this sampling distribution, making the target distribution in Eq.~\ref{eq:GC} no longer stationary and the approximation process difficult.
We alleviate this issue by ignoring $\pi_k$, effectively treating it as a distribution independent of $k$.
This transforms Eq.~\ref{eq:posterior} into:
\begin{equation}
   \hat{P}(M_k \mid s_t, a_t, s_{t+1}) \propto {P(M_k \mid s_t){\modprim(s_{t+1} \mid s_t, a_t, M_k)}}
  \label{eq:phat} 
\end{equation}

\subsection{Learning}
Since the focus of this work is on difficult continuous action problems, we mostly concentrate on the issue of policy optimization and how it integrates with the gating controller.
The standard policy (SP) optimization objective is:
\begin{equation}
  \underset{\theta}{\text{maximize }} \mathcal{L}^{SP} = \mathbb{E}_{\rho_0, \pi_{\theta}}[\pi_{\theta}(a_t \mid s_t) Q_{\pi_{\theta}}(s_t, a_t)]
\end{equation}
With baseline subtraction for variance reduction, this turns into~\cite{Schulman2015-nj}:
\begin{equation}
  \underset{\theta}{\text{maximize }} \mathcal{L}^{PG} = \mathbb{E}_{\rho_0, \pi_{\theta}}[\pi_{\theta}(a_t \mid s_t) \hat{A}_t]
\end{equation}
where $\hat{A}_t$ is an estimator of the advantage function~\cite{Baird1994-jl}.

In MPHRL, we directly use the mixture policy as defined by Eq.~\ref{eq:mix}.
The standard policy gradients (PG) get weighted by the probability outputs of the gating controller, enforcing the required specialization by factorizing into:
\begin{equation}
  \hat{g}_k = \mathbb{E}_{\rho_0, \pi_{\theta_k}}\left[P_{\phi}(M_k \mid s_t) \nabla_{\theta_k}\log \pi_{\theta_k}(a_t \mid s_t) \hat{A}_t\right]
\end{equation}

In practice, we use the Clipped PPO objective~\cite{ppo} instead to perform stable updates by limiting the step size.
This includes adding a baseline estimator (BL) parameterized by $\psi$ for value prediction and variance reduction. We optimize $\psi$ according to the following loss:
\begin{equation}
  \mathcal{L}^{BL} = \mathbb{E}\left[ \left\lVert V_{\psi} - V_{\pi_{\theta}} \right\rVert^2 \right]
\end{equation}
We summarize this single-task learning algorithm in Algorithm~\ref{algo:HRL-single}, which results in a set of decomposed subpolicies, $\pi_{\theta_1}, \ldots \pi_{\theta_K}$, and a gating controller $P_\phi$ that can modulate between them to solve the task under consideration.
\begin{algorithm}
\caption{MPHRL:\@ single-task learning}
\begin{algorithmic}[1]
  \STATE Initialize $P_{\phi}, \pi_{\theta} = \{ \pi_{\theta_1}, \ldots, \pi_{\theta_K} \}$, $V_{\psi}$
  \WHILE{not converged}{
    \STATE Rollout trajectories $\tau \sim \pi_{\theta,\phi}$
    \STATE Compute advantage estimates $\hat{A}_{\tau}$
    \STATE Optimize $\mathcal{L}^{PG}$ wrt $\theta_1, \ldots, \theta_K$ \\ \qquad with expectations taken over $\tau$
    \STATE Optimize $\mathcal{L}^{BL}$ wrt ${\psi}$ \\ \qquad with expectations taken over $\tau$
    \STATE Optimize $\mathcal{L}^{GC}$ wrt $\phi$ \\ \qquad with expectations taken over $\tau$
}
\ENDWHILE
\end{algorithmic}
\label{algo:HRL-single}
\end{algorithm}

\textit{Lifelong learning}: 
We have shown how MPHRL can decompose a single complex task solution into different functional components. 
Complex tasks often share structure and can be decomposed into similar sets of subtasks.
Different tasks however require different recomposition of similar subtasks.
Therefore, we transfer the subpolicies to learn target tasks, but not the gating controller or the baseline estimator.
We summarize the lifelong learning algorithm in Algorithm~\ref{algo:HRL-ll}, with the global variable \texttt{RESET} set to true.

\begin{algorithm}
\caption{MPHRL: lifelong learning}
\begin{algorithmic}[1]
  \STATE Initialize $P_{\phi}$, $\pi_{\theta} = \{ \pi_{\theta_1}, \ldots, \pi_{\theta_K} \}, V_{\psi}$
  \FOR{Tasks $(\mathcal{R}_i, \mathcal{T}_i) \sim D$}
  \IF {\texttt{RESET}}
  \STATE Initialize $P_{\phi}$, $V_{\psi}$
  \ENDIF
  \WHILE{not converged}{
    \STATE Rollout trajectories $\tau \sim \pi_{\theta,\phi}$
    \STATE Compute advantage estimates $\hat{A}_{\tau}$
    \STATE Optimize $\mathcal{L}^{PG}$ wrt $\theta_1, \ldots, \theta_K$ \\ \qquad with expectations taken over $\tau$ 
    \STATE Optimize $\mathcal{L}^{BL}$ wrt ${\psi}$ \\ \qquad with expectations taken over $\tau$   
    \STATE Optimize $\mathcal{L}^{GC}$ wrt $\phi$ \\ \qquad with expectations taken over $\tau$
  }
  \ENDWHILE
\ENDFOR
\end{algorithmic}
\label{algo:HRL-ll}
\end{algorithm}

\section{Experiments}

Our experiments aim to answer two questions:
(a) can model primitives ensure task decomposition?
(b) does such decomposition improve transfer for lifelong learning?

We evaluate our approach in two challenging domains: a MuJoCo~\cite{mujoco} ant navigating different mazes and a Stacker~\cite{dmcontrol} arm picking up and placing different boxes.
In our experiments, we use subpolicies that have Gaussian action distributions, with mean given by a multi-layer perceptron taking observations as input and standard deviations given by a different set of parameters.
MPHRL's gating controller outputs a categorical distribution and is parameterized by another multi-layer perceptron.
We also use a separate multi-layer perceptron for the baseline estimator.
We use the standard PPO algorithm as a baseline to compare against MPHRL.
Transferring network weights empirically led to worse performance for standard PPO. %
Hence, we re-initialize its weights for every task.
For fair comparison, we also shrink the hidden layer size of MPHRL's subpolicy networks from 64 to 16.
We conduct each experiment across 5 different seeds.
Error bars represent the standard deviation from the mean.

The focus of this work is on understanding the usefulness of model primitives for task decomposition and the resulting improvement in sample efficiency from transfer.
To conduct controlled experiments with interpretable results, we hand-designed model primitives using the true next state provided by the environment simulator.
Concretely, we apply distinct multivariate Gaussian noise models with covariance $\sigma\Sigma$ to the true next state. We then sample from this distribution to obtain the mean of the probability distribution of a model primitive's next state prediction, using $\Sigma$ as its covariance. Here, $\sigma$ is the noise scaling factor that distinguishes model primitives, while $\Sigma$ refers to the empirical covariance of the sampled next states:
\begin{align}
    \mu &\sim \mathcal{N}(s_{t+1}, \sigma_k\Sigma) \\
    \modprim{(s_{t+1} \mid s_t, a_t, M_k)} &= \mathcal{N}(\mu, \sigma_k\Sigma)
\end{align}

Using $\Sigma$ as opposed to a constant covariance is essential for controlled experiments because different elements of the observation space have different orders of magnitude.
Sampling $\mu$ from a distribution effectively adds random bias to the model primitive's next state probability distribution.

Hyperparameter details are in Table~\ref{tbl:hypermphrl}, and our code is freely available at \href{http://github.com/sisl/MPHRL}{http://github.com/sisl/MPHRL}.
\begin{table}[h!]
    \caption{Hyperparameters: MPHRL and baseline PPO}
    \resizebox{\linewidth}{!}{
    \begin{tabular}{llc}
    \toprule
    Category & Hyper-parameter & Value \\ \midrule
    Num. model primitives: & L-Maze & 2 \\
    Maze & D-Maze &  4 \\
     & Standard 10-Maze& 4 \\
    & H-V Corridors & 2 \\
    & Velocity&  2 \\
         & Extra&  5 \\\midrule
    Num. model primitives: & Standard 8-P\&P & 12 \\
    8-P\&P\tablefootnote{8-Pickup\&Place.}& Box-only &  2 \\
     & Action-only & 6 \\\midrule
    Gating controller: & Hidden layers & 2 \\
    Network & Hidden dimension & 64 \\\midrule 
    Gating controller: & Single / Source\tablefootnote{Single task refers to L-Maze and D-Maze; source and target tasks refer to the first task and all subsequent tasks in a lifelong learning taskset, respectively.} (Maze) & $1\times10^{-3}$ \\ 
    Base learning rate & Single / Source (8-P\&P) & $3\times10^{-2}$ \\ 
     & Target tasks & $3\times10^{-3}$ \\ 
         \midrule
    Gating controller: & Single / Source & 1 \\ 
    Num. epoches / batch & Target tasks & 10 \\\midrule
    Baseline and model & Hidden layers & 2 \\
    primitive networks\tablefootnote{Baseline network hyperparameters apply to both MPHRL and baseline PPO; model primitive networks are for experiments with learned model primitives only.} & Hidden dimension & 64 \\ 
         & Base learning rate & $3\times10^{-4}$\\ \midrule
    \multirow{4}{*}{Subpolicy networks\tablefootnote{The baseline PPO has no subpolicies, so the subpolicy network \textit{is} the policy network.}} & Hidden layers & 2 \\
         & Hidden dimension (MPHRL) & 16 \\ 
         & Hidden dimension (PPO) & 64 \\ 
         & Base learning rate & $3\times10^{-4}$\\ \midrule
    \multirow{13}{*}{Optimization} & Num. actors (Maze) & 16 \\
         & Num. actors (8-P\&P) & 24 \\
         & Batch size / actor (Maze) & 2048 \\
         & Batch size / actor (8-P\&P) & 1536 \\
         & Max. timesteps / task & $3 \times 10^7$ \\
         & Minibatch size / actor & 256 \\
         & Num. epoches / batch\tablefootnote{Baseline and subpolicy networks only.} & 10 \\
         & Discount ($\gamma$) & 0.99 \\
         & GAE parameter ($\lambda$) & 0.95 \\
         & PPO clipping coeff. ($\epsilon$) & 0.2 \\
         & Gradient clipping & None \\
         & VF coeff. ($c_1$) & 1.0 \\
         & Entropy coeff. ($c_2$) & 0 \\
         & Optimizer & Adam \\
         \bottomrule
    \end{tabular}
    }
    \label{tbl:hypermphrl}
\end{table}

\subsection{Single-task Learning}
\begin{figure}
  \centering
  \captionsetup{justification=centering}
  \begin{subfigure}[b]{0.5\columnwidth}
    \centering
    \begin{minipage}[b]{\columnwidth}
      \centering
      \includegraphics[width=0.5\columnwidth]{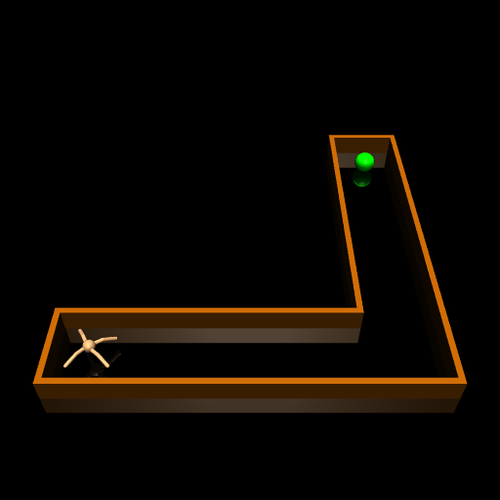}
    \end{minipage}\\[\baselineskip]
    \begin{minipage}[b]{\columnwidth}
      \centering
      \includegraphics[width=0.5\columnwidth]{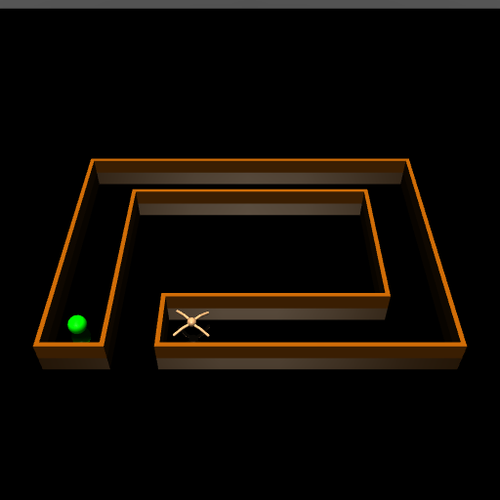}
    \end{minipage}
    \caption{L-Maze (top) and D-Maze (bottom)}\label{LDAnt}
  \end{subfigure}
  \begin{subfigure}[b]{0.49\columnwidth}
    \centering
    \includegraphics[width=\columnwidth]{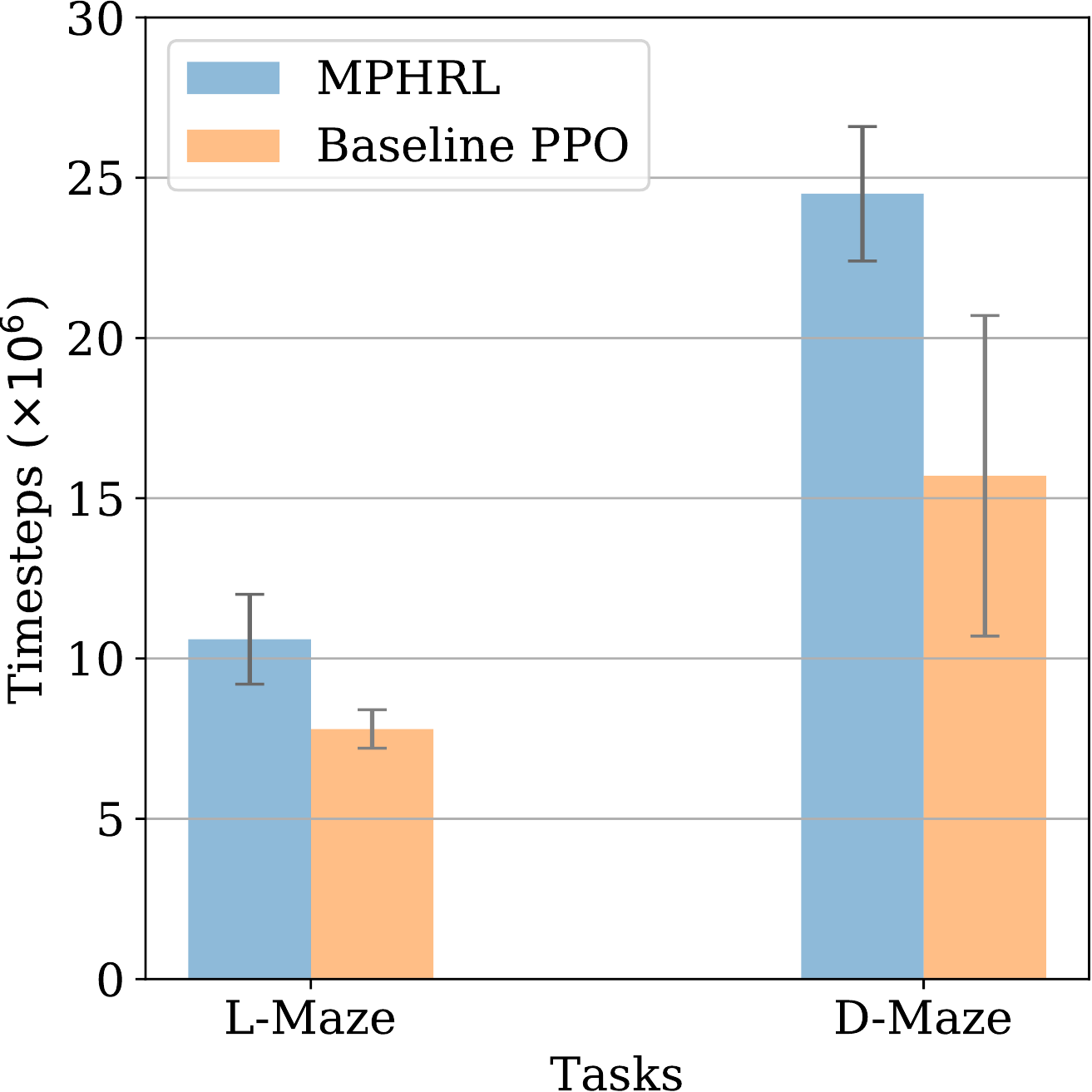}
   \caption{Performance}\label{tbl:single}
  \end{subfigure}
  \caption{Single-task learning}
\end{figure}

First, we focus on two single-task learning experiments where MPHRL learns a number of interpretable subpolicies to solve a single task.
Both the L-Maze and D-Maze (Figure~\ref{LDAnt}) tasks require the ant to learn to walk and reach the green goal within a finite horizon.
For both tasks, both the goal and the initial ant locations are fixed.
For the L-Maze, the agent has access to two model primitives, one specializing in the horizontal (E, W) corridor and the other specializing in the vertical (N, S) corridor of the maze.
Similarly for the D-Maze, the agent has access to four model primitives, one specializing in each N, S, E, W corridor of the maze.
In their specialized corridors, the noise scaling factor $\sigma = 0$. 
Outside of their regions of specialization, $\sigma = 0.5$.
The observation space includes the standard joint angles and positions, lidar information tracking distances from walls on each side, and the Manhattan distance to the goal.
Figure~\ref{tbl:single} shows the experimental results on these environments.
Notice that using model primitives can make the learning problem more difficult and increase the sample complexity on a single task.
This is expected, since we are forcing the agent to decompose the solution, which could be unnecessary for easy tasks.
However, we will observe in the following section that this decomposition can lead to remarkable improvements in transfer performance during lifelong learning.

\subsection{Lifelong Learning}
To evaluate our framework's performance at lifelong learning, we introduce two tasksets.
\subsubsection{10-Maze}
\begin{figure}
    \centering
    \begin{subfigure}[b]{0.49\columnwidth}
    \centering
    \includegraphics[width=\columnwidth]{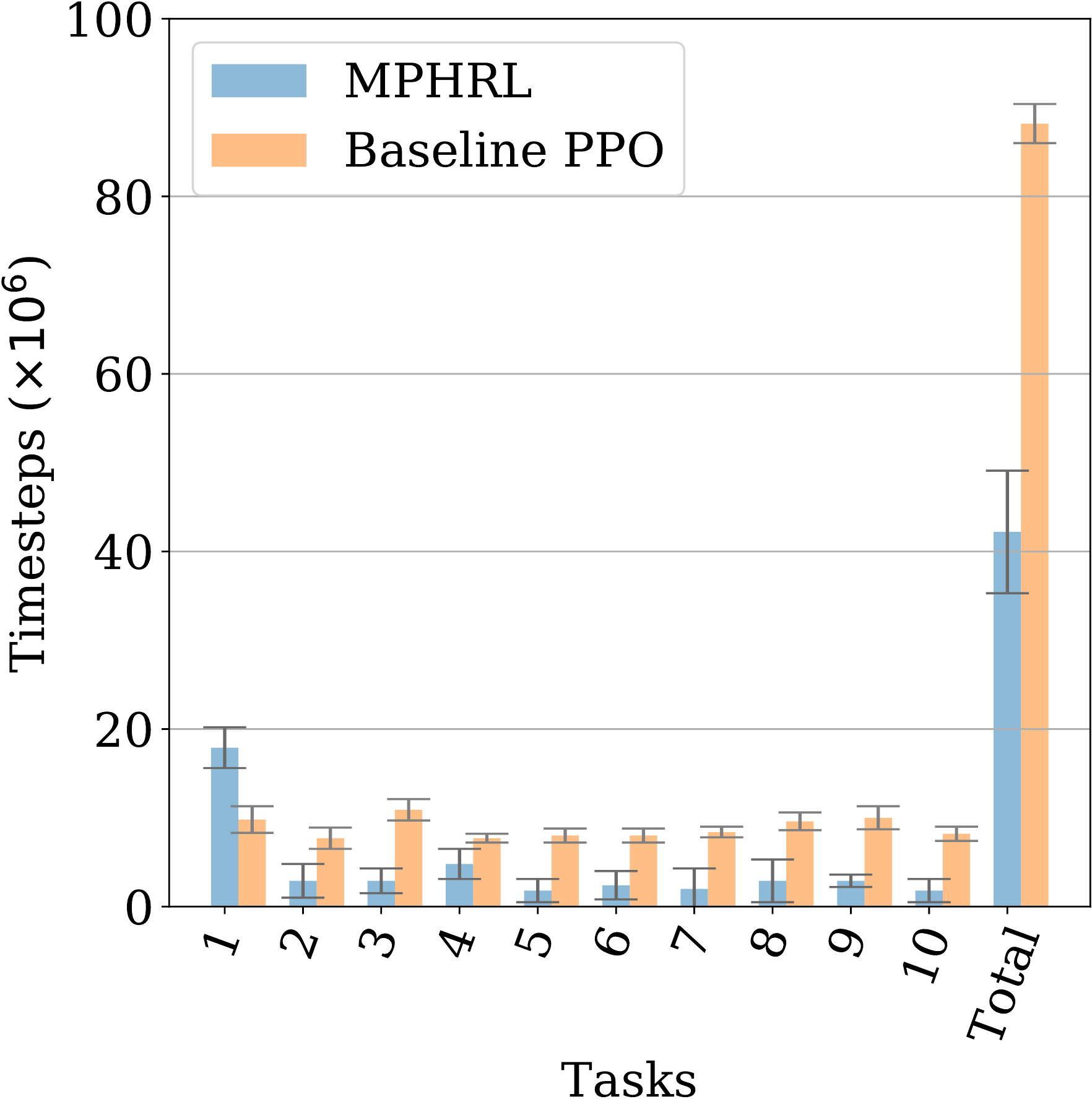}\caption{10-Maze}\label{fig:ll10}
    \end{subfigure}
    \begin{subfigure}[b]{0.49\columnwidth}
    \centering
    \includegraphics[width=\columnwidth]{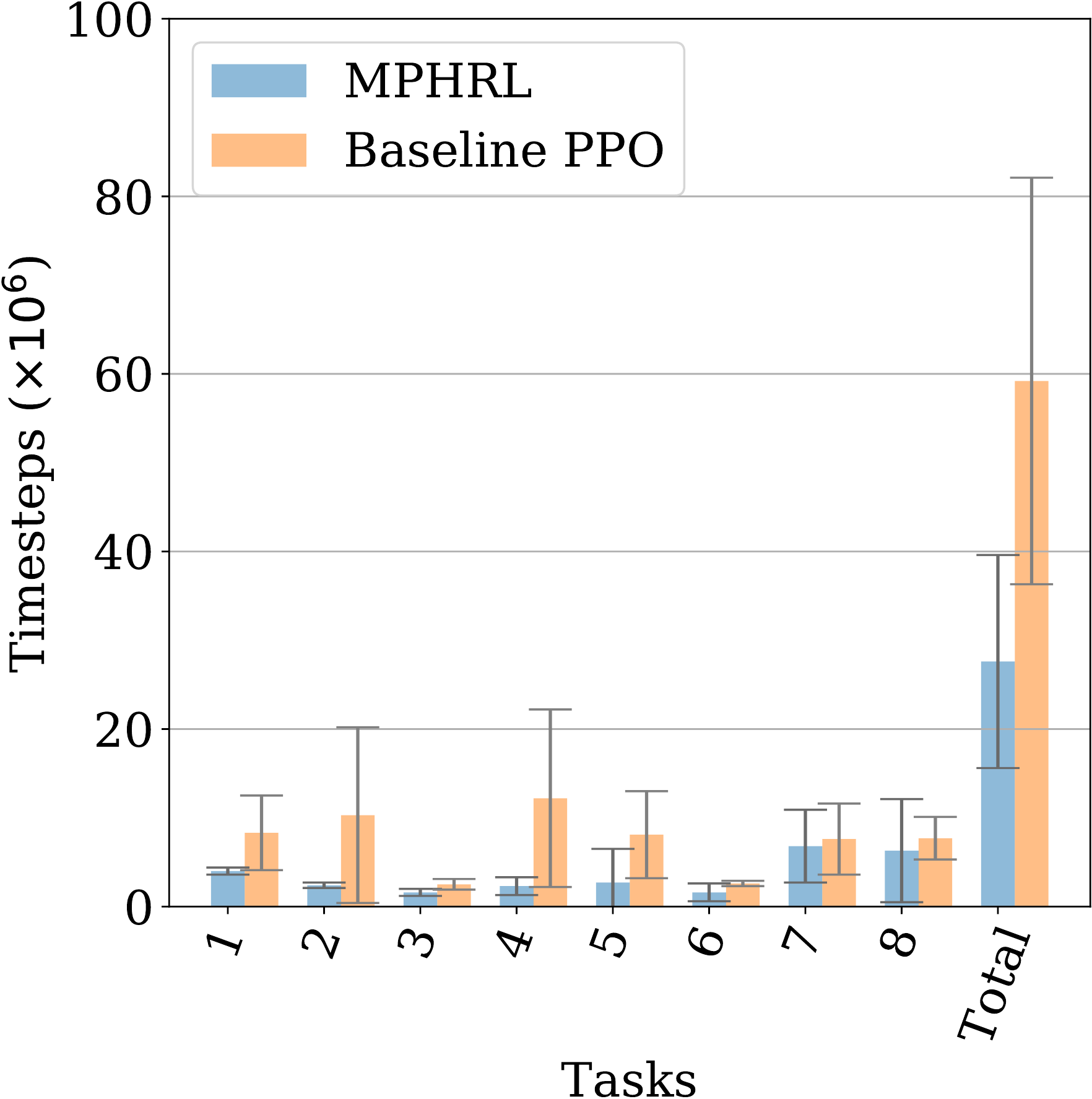}\caption{8-Pickup\&Place}\label{fig:8pp}
    \end{subfigure}
  \caption{MPHRL vs. PPO for lifelong learning}
  \label{fig:ll}
\end{figure}

\begin{figure*}
  \begin{center}
    \begin{subfigure}[t]{0.095\textwidth}\includegraphics[width=\linewidth]{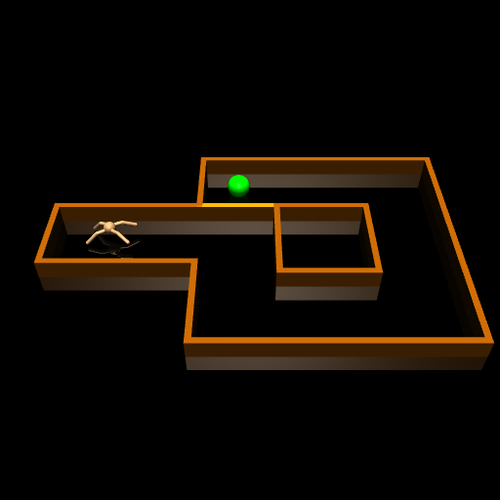}\caption{}\end{subfigure}
    \begin{subfigure}[t]{0.095\textwidth}\includegraphics[width=\linewidth]{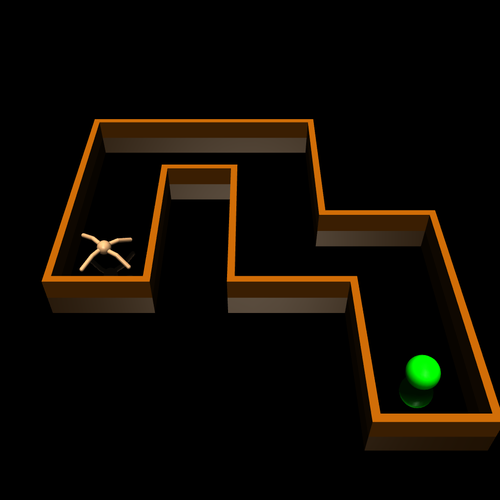}\caption{}\end{subfigure}
    \begin{subfigure}[t]{0.095\textwidth}\includegraphics[width=\linewidth]{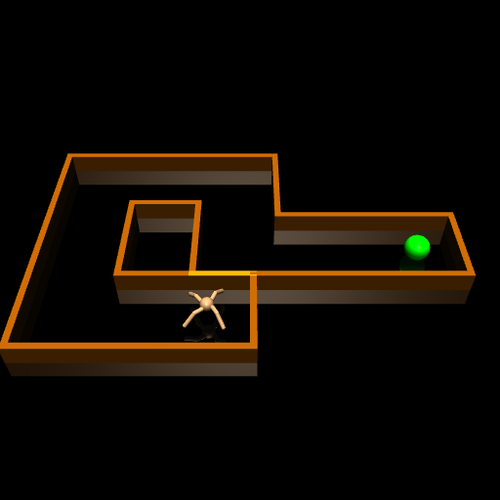}\caption{}\end{subfigure}
    \begin{subfigure}[t]{0.095\textwidth}\includegraphics[width=\linewidth]{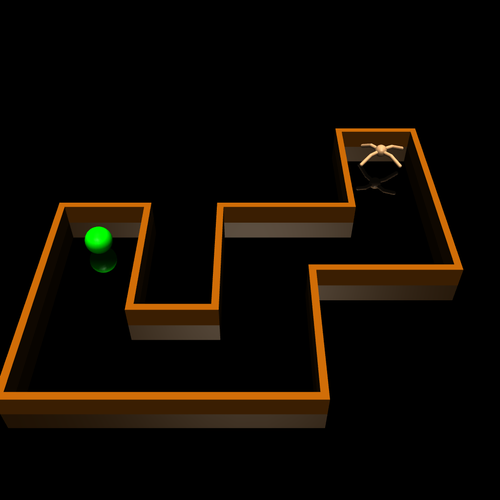}\caption{}\end{subfigure}
    \begin{subfigure}[t]{0.095\textwidth}\includegraphics[width=\linewidth]{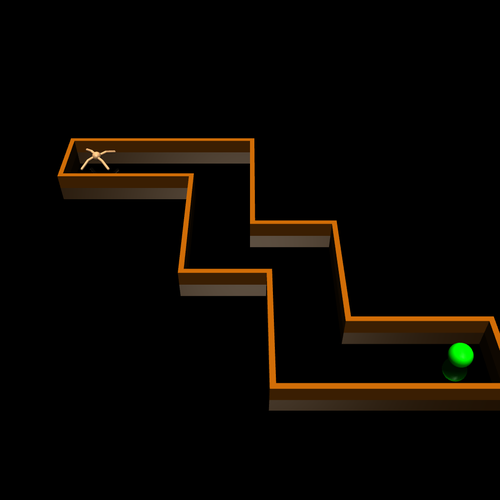}\caption{}\end{subfigure}
    \begin{subfigure}[t]{0.095\textwidth}\includegraphics[width=\linewidth]{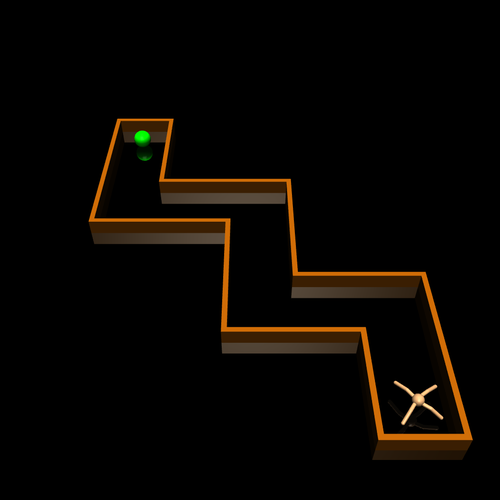}\caption{}\end{subfigure}
    \begin{subfigure}[t]{0.095\textwidth}\includegraphics[width=\linewidth]{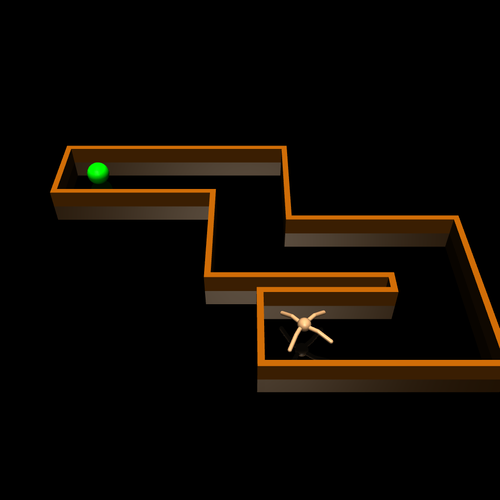}\caption{}\end{subfigure}
    \begin{subfigure}[t]{0.095\textwidth}\includegraphics[width=\linewidth]{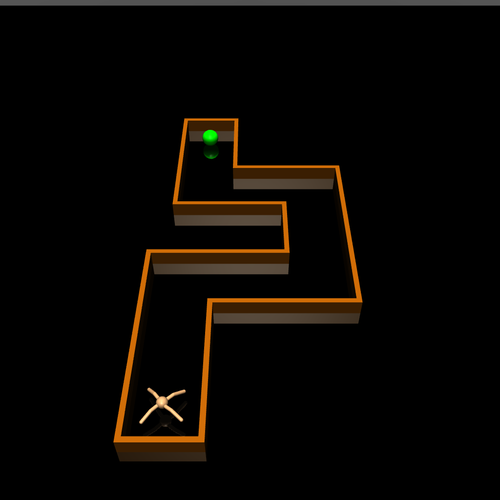}\caption{}\end{subfigure}
    \begin{subfigure}[t]{0.095\textwidth}\includegraphics[width=\linewidth]{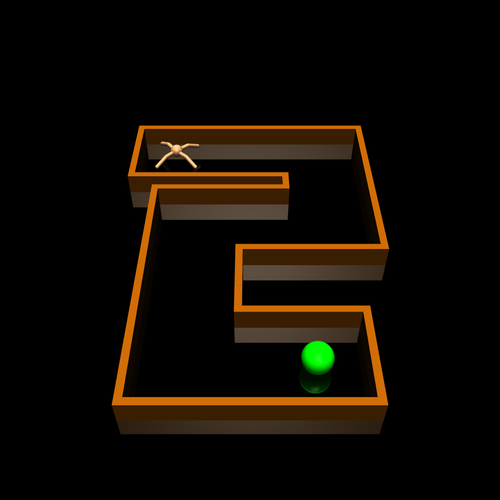}\caption{}\end{subfigure}
    \begin{subfigure}[t]{0.095\textwidth}\includegraphics[width=\linewidth]{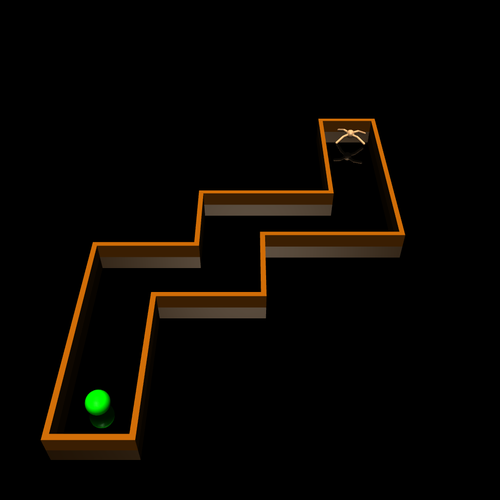}\caption{}\end{subfigure}
  \end{center}
  \caption{10-Maze lifelong learning taskset}\label{10Maze}
\end{figure*}

To evaluate MPHRL's performance in lifelong learning, we generate a family of 10 random mazes for the MuJoCo Ant environment, referred to as the 10-Maze taskset (Figure \ref{10Maze}) hereafter.
The goal, the observation space, the Gaussian noise models, and the model primitives remain the same as in D-Maze.
The agent has a maximum of $3 \times 10^7$ timesteps to reach 80\% success rate in each of the 10 tasks.
As shown in Figure~\ref{fig:ll10}, MPHRL requires nearly double the number of timesteps to learn the decomposed subpolicies in the first task.
However, this cost gets heavily amortized over the entire taskset, with MPHRL taking half the total number of timesteps of the baseline PPO, exhibiting strong subpolicy transfer.

\begin{figure*}
  \centering
  \captionsetup{justification=centering}
    \begin{subfigure}[t]{0.12\textwidth}\centering\includegraphics[width=\linewidth]{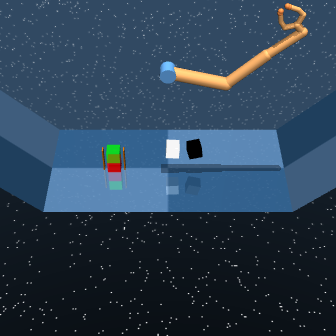}\caption{\\B1 $\rightarrow$ T1\\ B2 $\rightarrow$ T2}\end{subfigure}
    \begin{subfigure}[t]{0.12\textwidth}\centering\includegraphics[width=\linewidth]{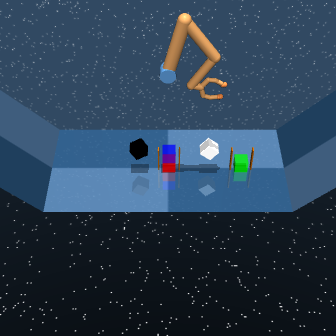}\caption{\\B2 $\rightarrow$ T1\\ B1 $\rightarrow$ T2\\ B1 $\rightarrow$ T3}\end{subfigure}
    \begin{subfigure}[t]{0.12\textwidth}\centering\includegraphics[width=\linewidth]{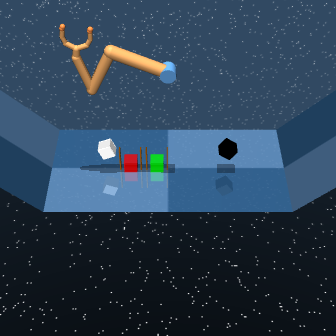}\caption{\\B2 $\rightarrow$ T1\\ B2 $\rightarrow$ T2}\end{subfigure}
    \begin{subfigure}[t]{0.12\textwidth}\centering\includegraphics[width=\linewidth]{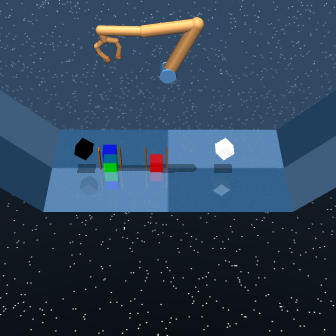}\caption{\\B1 $\rightarrow$ T1\\ B1 $\rightarrow$ T2\\ B2 $\rightarrow$ T3}\end{subfigure}
    \begin{subfigure}[t]{0.12\textwidth}\centering\includegraphics[width=\linewidth]{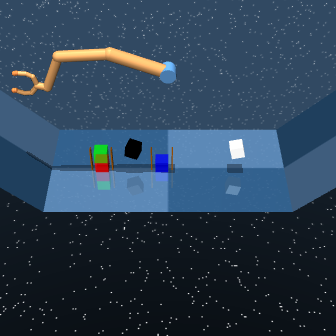}\caption{\\B1 $\rightarrow$ T1 \\ B2 $\rightarrow$ T2 \\ B2 $\rightarrow$ T3}\end{subfigure}
    \begin{subfigure}[t]{0.12\textwidth}\centering\includegraphics[width=\linewidth]{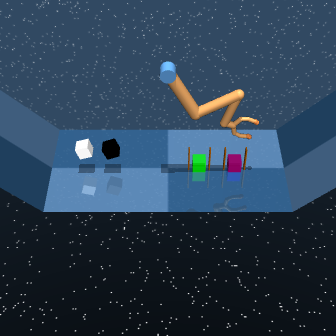}\caption{\\B1 $\rightarrow$ T1 \\B1 $\rightarrow$ T2 \\B1 $\rightarrow$ T3}\end{subfigure}
    \begin{subfigure}[t]{0.12\textwidth}\centering\includegraphics[width=\linewidth]{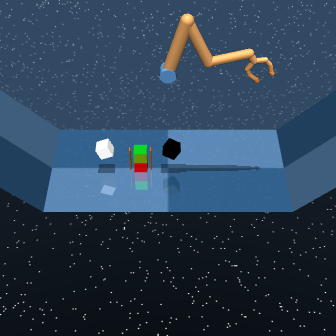}\caption{\\B2 $\rightarrow$ T1 \\B1 $\rightarrow$ T2}\end{subfigure}
    \begin{subfigure}[t]{0.12\textwidth}\centering\includegraphics[width=\linewidth]{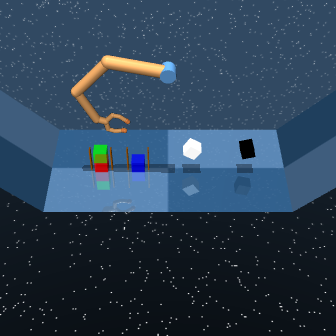}\caption{\\B2 $\rightarrow$ T1 \\B1 $\rightarrow$ T2 \\B1 $\rightarrow$ T3}\end{subfigure}
  \caption{8-Pickup\&Place lifelong learning taskset. B1 and B2 refer to Box1 (black) and Box2 (white); T1, T2, and T3 refer to Target 1 (red), Target 2 (green), and Target 3 (blue)}\label{stacker}
\end{figure*}

\begin{figure*}
    \centering
    \captionsetup{justification=centering}
    \begin{subfigure}[t]{0.33\textwidth}
    \centering
    \includegraphics[width=\columnwidth]{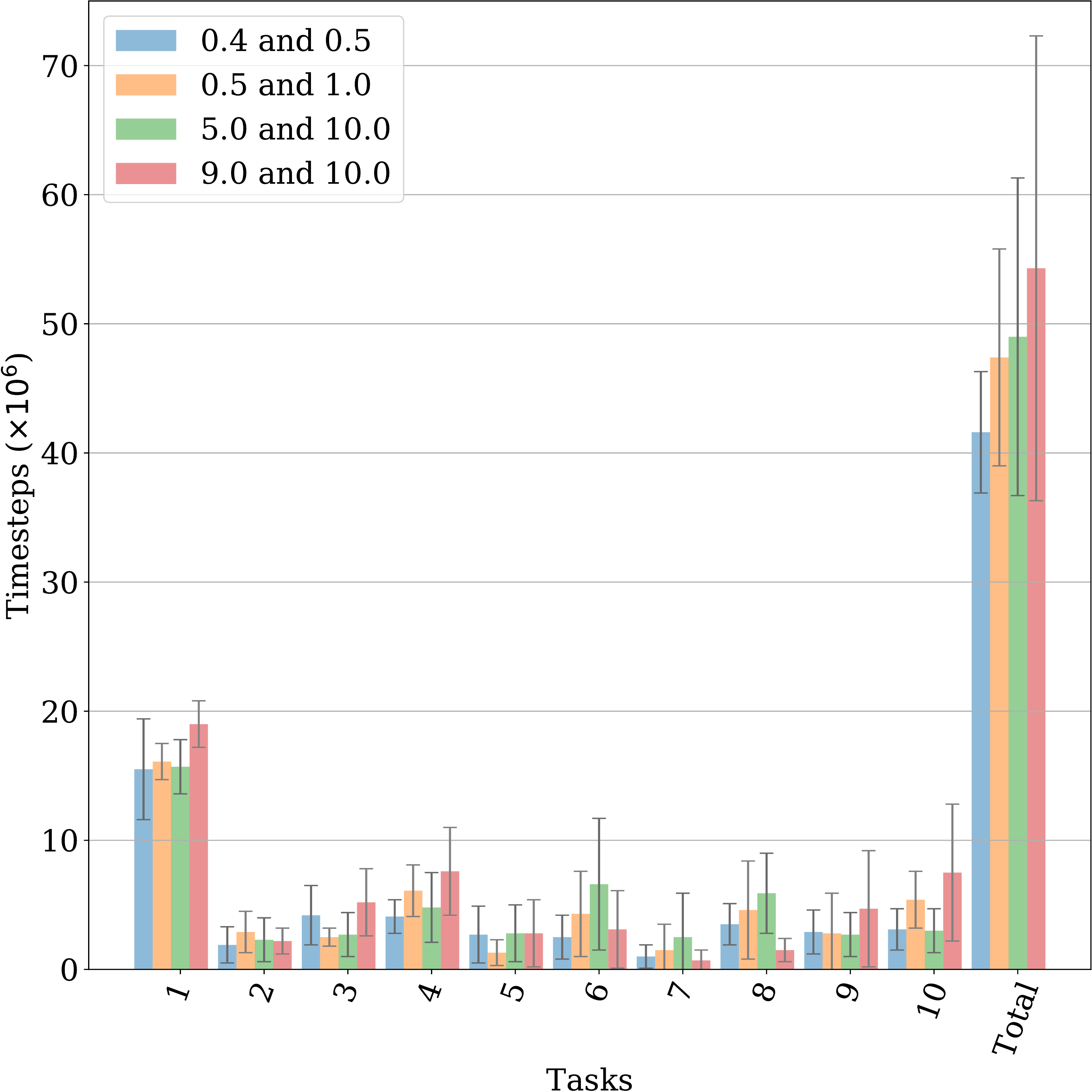}\caption{Effect of noisy model primitives}\label{fig:noise}
    \end{subfigure}
    \begin{subfigure}[t]{0.33\textwidth}
    \centering
    \includegraphics[width=\columnwidth]{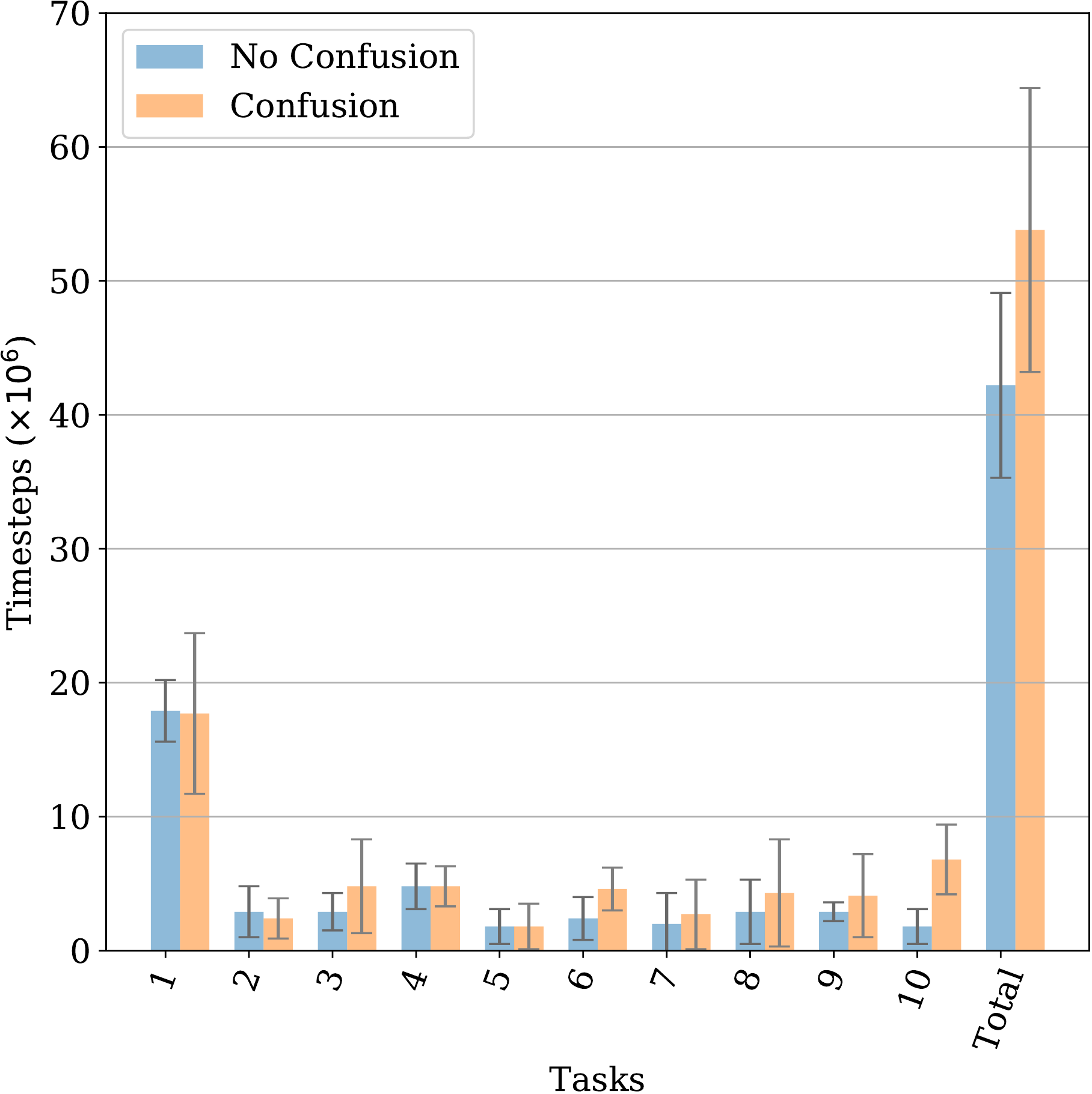}\caption{Effect of model primitive confusion at the corners}\label{fig:corner}
    \end{subfigure}
    \begin{subfigure}[t]{0.33\textwidth}
    \centering
    \includegraphics[width=\columnwidth]{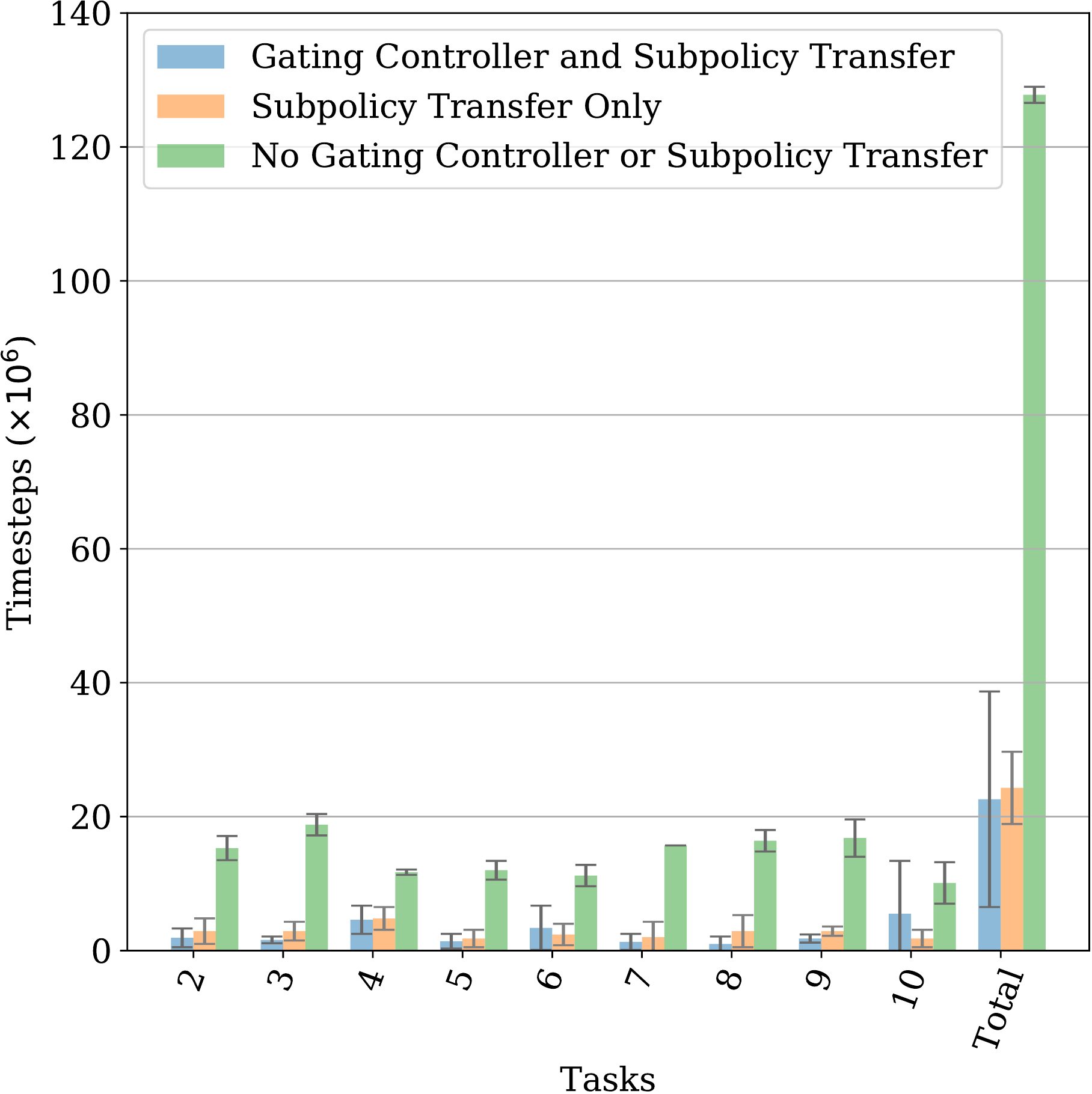}\caption{Effect of gating controller and subpolicy transfer (target tasks only)}\label{fig:transfer}
    \end{subfigure}
  \caption{10-Maze: MPHRL ablation}
  \label{fig:llablation}
\end{figure*}

\begin{table*}
  \centering
    \caption{Effect of suboptimal model primitive types (N/A indicates failure to solve the task within $3 \times 10^7$ timesteps)\label{tbl:diversity}}
    \resizebox{\linewidth}{!}{\begin{tabular*}{1.28\textwidth}{lr@{}S@{}S@{}S@{}S@{}S@{}S@{}S@{}S@{}S@{}S@{}S@{}S@{}S}
      \toprule
       & & \multicolumn{10}{c}{Timesteps to reach target average success rate of $80\%$ (10-Maze) and $75\%$ (8-Pickup\&Place) ($\times 10^6$)} \\ \cmidrule(l){3-13}
       Taskset & Model Primitives & 1 & 2 & 3 & 4 & 5 & 6 & 7 & 8 & 9 & 10 & Total\\ \midrule
       10-Maze & Extra & 21.8 \pm 1.5& 4.3 \pm 0.8& 3.5 \pm 0.9& 5.2 \pm 1.2& 0.7 \pm 0.5& 3.6 \pm 1.8& 1.2 \pm 1.1& 1.5 \pm 0.9& 3.1 \pm 0.7& 5.8 \pm 3.3& 50.7 \pm 4.3 \\
       10-Maze & H-V Corridors & 28.1 \pm 4.2& 2.4 \pm 0.6& 18.9 \pm 11.0& 28.8 \pm 2.8 & N/A &  &  &  &  &  &  \\
       10-Maze & Velocity  & 14.7 \pm 2.4& 1.4 \pm 1.2& 17.9 \pm 11.2& 20.5 \pm 13.2 & N/A &  &  &  &  &  & \\
       8-Pickup\&Place & Action-only & 5.1 \pm 0.6& 4.8 \pm 1.8& 16.8 \pm 12.6& 26.9 \pm 6.9 & N/A &  &  &  &  &  & \\
       8-Pickup\&Place & Box-only & 18.6 \pm 10.9 & N/A & &  &  &  &  &  &  &  & \\
      \bottomrule
    \end{tabular*}}
\end{table*}

\begin{table*}
  \centering
    \caption{10-Maze: effect of experience\label{tbl:exp}}
    \begin{tabular}{lS@{}S@{}S@{}S@{}S@{}S@{}S@{}S@{}S}
      \toprule
       & \multicolumn{9}{c}{Timesteps to reach $80\%$ average success rate ($\times 10^6$)} \\ \cmidrule(l){2-10}
       Experience & 1 & 2 & 3 & 4 & 5 & 6 & 7 & 8 & 9 \\ \midrule
      10 tasks & 0.6 \pm 0.1& 0.4 \pm 0.0& 0.6 \pm 0.0& 0.5 \pm 0.1& 0.4 \pm 0.0& 2.3 \pm 0.7& 0.7 \pm 0.2& 2.6 \pm 0.3& 0.7 \pm 0.1\\
      6 tasks & 3.1 \pm 0.4& 2.0 \pm 0.2& 3.5 \pm 0.7& 2.2 \pm 0.7& 1.8 \pm 0.3&  &\\
      \bottomrule
    \end{tabular}
\end{table*}
\subsubsection{8-Pickup\&Place}
We modify the Stacker task~\cite{dmcontrol} to create the 8-Pickup\&Place taskset.
As shown in Figure~\ref{stacker}, a robotic arm is tasked to bring 2 boxes to their respective goal locations in a certain order. 
Marked by colors red, green, and blue, the goal locations reside within two short walls forming a ``stack''.

Each of the 8 tasks has a maximum of 3 goal locations.
The observation space of the agent includes joint angles and positions, box and goal locations, their relative distances to each other, and the current stage of the task encoded as one-hot vectors.
The agent has access to six model primitives for each box that specialize in reaching above, lowering to, grasping, picking up, carrying, and dropping a certain box.
Similar to 10-Maze, model primitives have $\sigma$ of 0 within their specialized stages and $\sigma$ of 0.5 otherwise.
Figure~\ref{fig:8pp} shows MPHRL's experimental performance by learning twelve useful subpolicies for this taskset.
We notice again the strong transfer performance due to the decomposition forced by the model primitives.
Note that this taskset is much more complex than 10-Maze such that MPHRL even accelerates the learning of the first task.

\subsection{Ablation}
We conduct ablation experiments to answer the following questions:
\begin{enumerate}[leftmargin=2\parindent]
\item How much gain in sample efficiency is achieved by transferring subpolicies?
\item Can MPHRL learn the task decomposition even when the model primitives are quite noisy or when the source task does not cover all ``cases''?
\item When does MPHRL fail to decompose the solution?
\item What kind of diversity in the model primitives is essential for performance?
\item When does MPHRL lead to negative transfer?
\item Is MPHRL's gain in sample efficiency a result of hand-crafted model primitives and how does it perform with actual learned model primitives?
\end{enumerate}

\subsubsection{Model Noise}
MPHRL has the ability to decompose the solution even given bad model primitives.
Since the learning is done model-free, these suboptimal model primitives should not strongly affect the learning performance so long as they remain sufficiently distinct.
To investigate the limitations to this claim, we conduct five experiments using various sets of noisy model primitives.
Below, the first value corresponds to the noise scaling factor $\sigma$ within their individual regions of specialization, while the second value corresponds to $\sigma$ outside of their regions of specialization.
\begin{enumerate}[label=(\alph*),leftmargin=2\parindent]
    \item 0.4 and 0.5: good models with limited distinction
    \item 0.5 and 1.0: good models with reasonable distinction
    \item 5.0 and 10.0: bad models with reasonable distinction
    \item 9.0 and 10.0: bad models with limited distinction
    \item 0.5 and 0.5: good models with no distinction
\end{enumerate}
Shown in Figure~\ref{fig:noise}, while (a), (b), (c), and (d) exhibit limited degradation in performance, (d) experiences the most performance degradation on average.
On the other hand, in (e) MPHRL took $22.0\pm4.6$ million timesteps to solve the first task and $2.8\pm1.6$ million timesteps to solve the second task, but failed to solve the third task within 30 million timesteps. This is because the model primitives are identical and provide no information about task decomposition.
In summary, MPHRL is robust against bad model primitives so long as the they maintain some relative distinction.
Similar observations hold true for the 8-Pickup\&Place taskset where noise models with distinctive models with large noise of $\sigma = 5$ and $\sigma = 20$ show little deterioration in performance, taking $15.8 \pm 5.5$ million timesteps to reach $75\%$ average success rate.%

\subsubsection{Overlapping Model Primitives}
We next test the condition when there is substantial overlap in regions of specialization between different model primitives.
For the 10-Maze taskset, the most plausible region for this confusion is at the corners.
In this experiment, within each corner, the two model primitives whose specialized corridors share the corner have $\sigma = 0$ while the other two have $\sigma = 0.5$.
Figure~\ref{fig:corner} shows the performance for model primitive confusion against the standard set of model primitives with no confusion.
We observe that despite some performance degradation, MPHRL continues to outperform the PPO baseline.

\subsubsection{Model Diversity}
Having tested MPHRL against noises, we experimented with undesirable model primitives for 10-Maze:
\begin{enumerate}[label=(\alph*),leftmargin=2\parindent]
    \item \textit{Extra}: a fifth model primitive that specializes in states where the ant is moving horizontally;
    \item \textit{H-V corridors}: $2$ model primitives specializing in horizontal (E, W) and vertical (N, S) corridors respectively;
    \item \textit{Velocity}: 2 model primitives specializing in states where the ant is moving horizontally or vertically;
\end{enumerate}
and for 8-Pickup\&Place:
\begin{enumerate}[label=(\alph*),leftmargin=2\parindent]
    \item \textit{Box-only}: 2 model primitives for all actions on 2 boxes;
    \item \textit{Action-only}: 6 model primitives for 6 actions performed on boxes: reach above, lower to, grasp, pick up, carry, and drop.
\end{enumerate}

Table~\ref{tbl:diversity} shows MPHRL is susceptible to performance degradation given undesirable sets of model primitives. 
However, MPHRL still outperforms baseline PPO when given an extra, undesirable model primitive.
This indicates that for best transfer, the model primitives need to approximately capture the structure present in the taskset.

\begin{figure}
    \centering
    \includegraphics[width=.8\columnwidth]{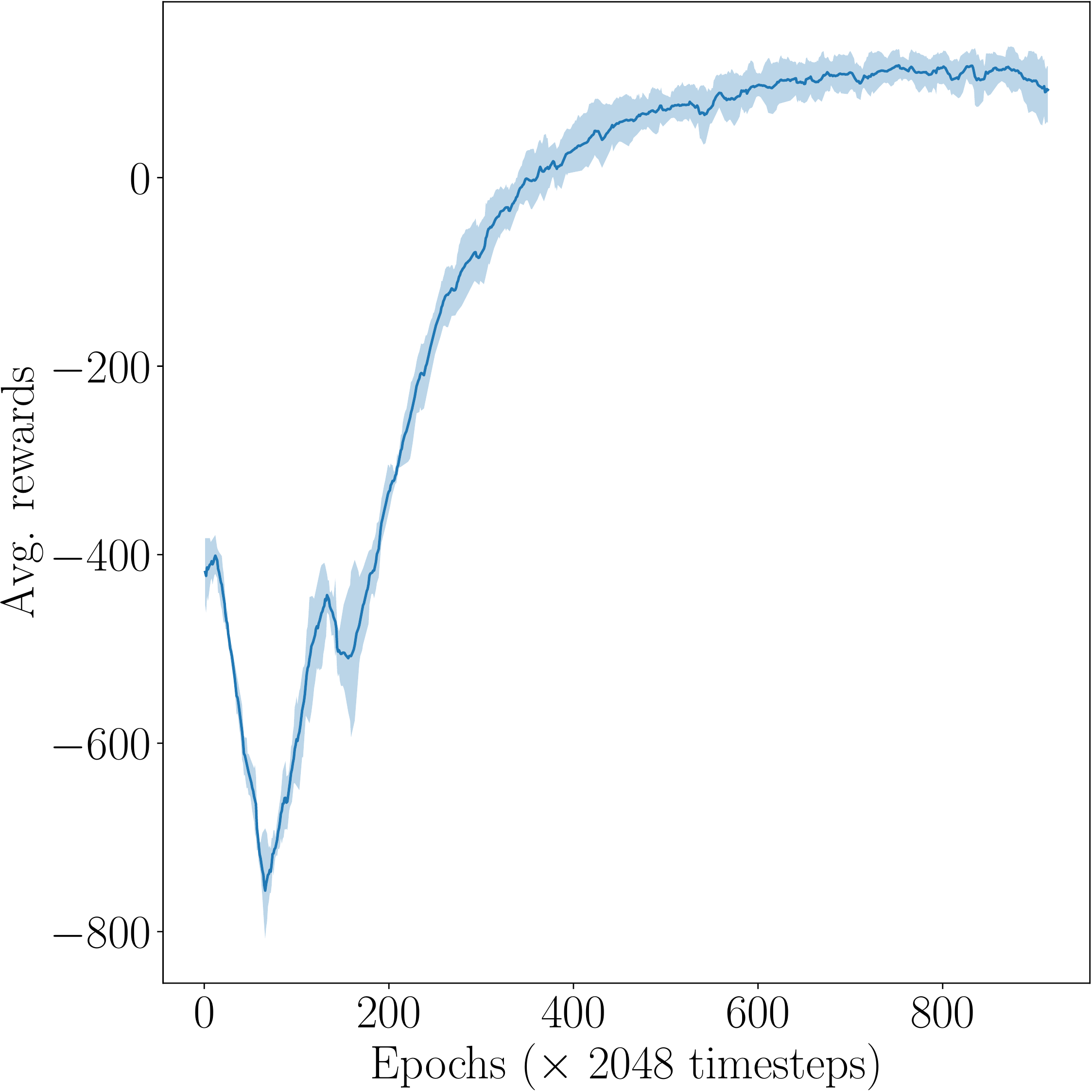}
    \caption{Average rewards of MPHRL when using an oracle gating controller. The reward threshold for reaching 80\% success rate for the first task is approximately 800.}
    \label{fig:oracle}
\end{figure}
\begin{figure}
    \centering
    \includegraphics[width=0.85\columnwidth]{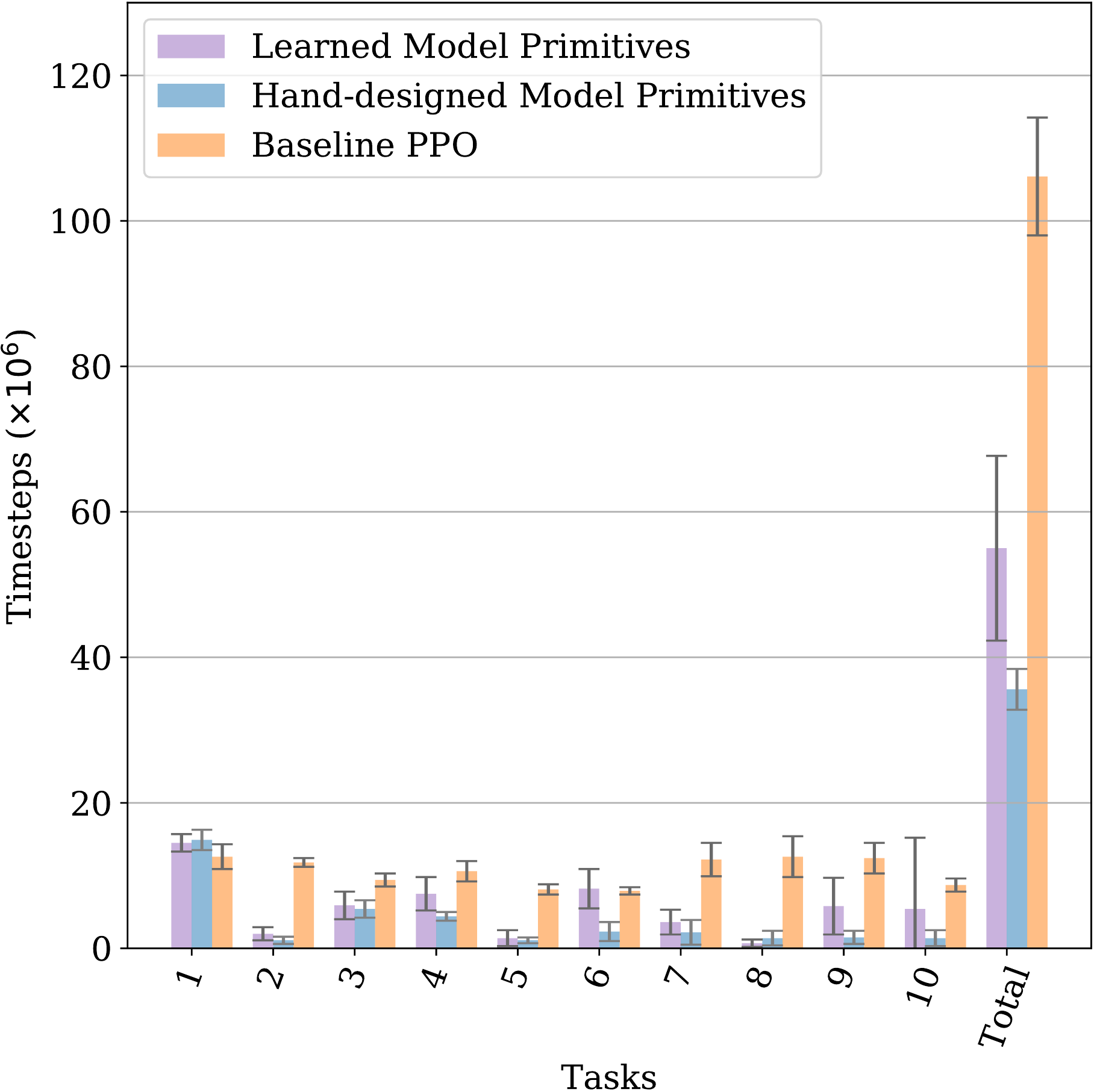}
    \caption{10-Maze-v2: partial decomposition and learned model primitives. Success threshold is at $70\%$.}
    \label{fig:learnmp}
\end{figure}

\subsubsection{Negative Transfer and Catastrophic Forgetting}
Lifelong learning agents with neural network function approximators face the problem of negative transfer and catastrophic forgetting.
Ideally, they should find the solution quickly if the task has already been seen. 
More generally, given two sets of tasks $T$ and $T'$ such that $T \subset T'$, after being exposed to $T'$ the agent should perform no worse, and preferably better, than had it been exposed to $T$ only.

In this experiment, we restore the subpolicy checkpoints after solving the $10$ tasks and evaluate MPHRL's learning performance for the first 9 tasks.
Similarly, we restore the subpolicy checkpoints after solving $6$ tasks and evaluate MPHRL's performance on the first 5 tasks.
The gating controller is reset for each task as in earlier experiments.
We summarize the results in Table~\ref{tbl:exp}.
Subpolicies trained sequentially on 6 or 10 tasks quickly relearn the required behavior for all previously seen tasks, implying no catastrophic forgetting.
Moreover, if we compare the 10-task result to the 6-task result, we see remarkable improvements at transfer. 
This implies negative transfer is limited with this approach.

\begin{figure*}
  \begin{center}
    \begin{subfigure}[t]{0.095\textwidth}\includegraphics[width=\linewidth]{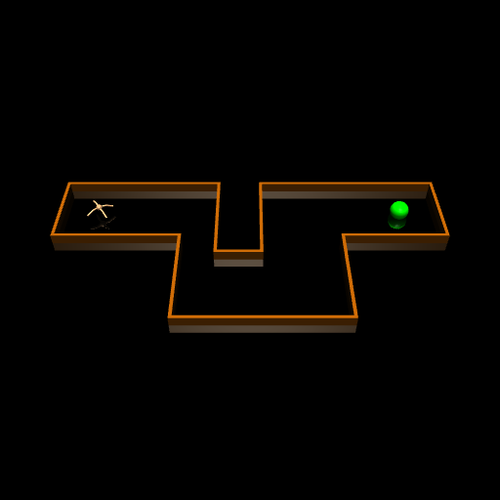}\end{subfigure}
    \begin{subfigure}[t]{0.095\textwidth}\includegraphics[width=\linewidth]{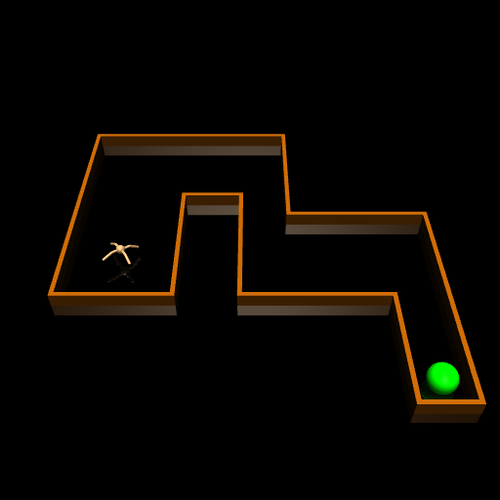}\end{subfigure}
    \begin{subfigure}[t]{0.095\textwidth}\includegraphics[width=\linewidth]{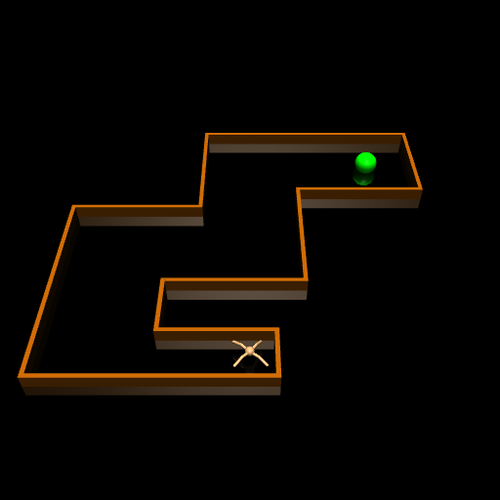}\end{subfigure}
    \begin{subfigure}[t]{0.095\textwidth}\includegraphics[width=\linewidth]{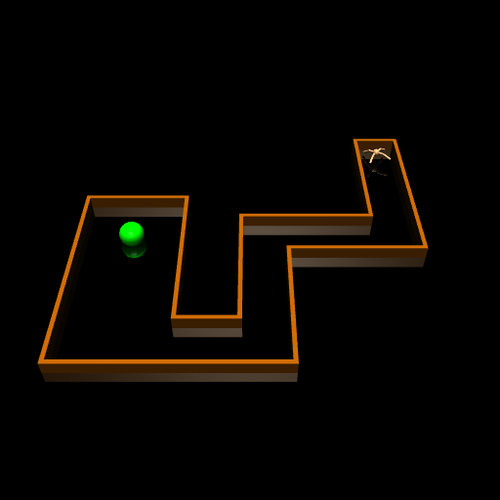}\end{subfigure}
    \begin{subfigure}[t]{0.095\textwidth}\includegraphics[width=\linewidth]{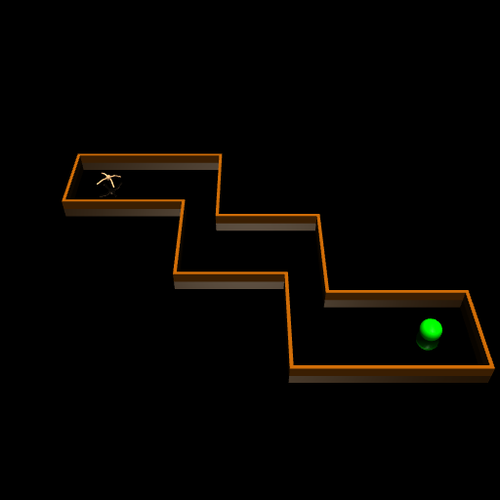}\end{subfigure}
    \begin{subfigure}[t]{0.095\textwidth}\includegraphics[width=\linewidth]{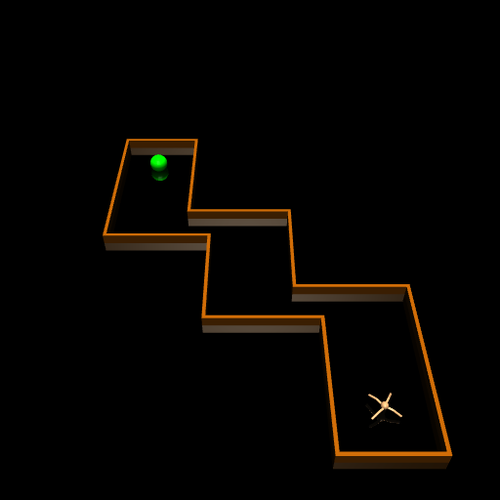}\end{subfigure}
    \begin{subfigure}[t]{0.095\textwidth}\includegraphics[width=\linewidth]{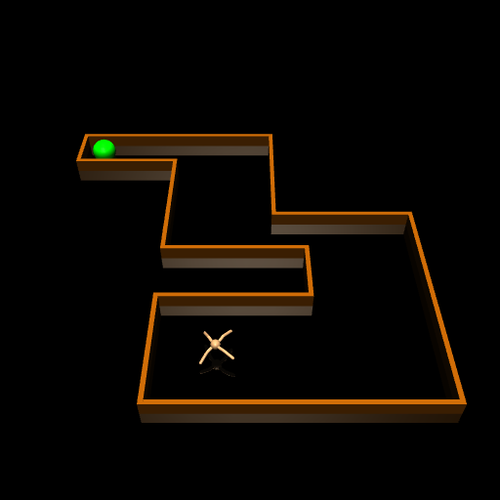}\end{subfigure}
    \begin{subfigure}[t]{0.095\textwidth}\includegraphics[width=\linewidth]{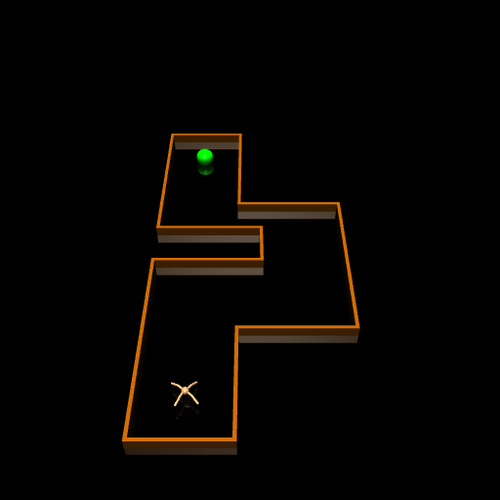}\end{subfigure}
    \begin{subfigure}[t]{0.095\textwidth}\includegraphics[width=\linewidth]{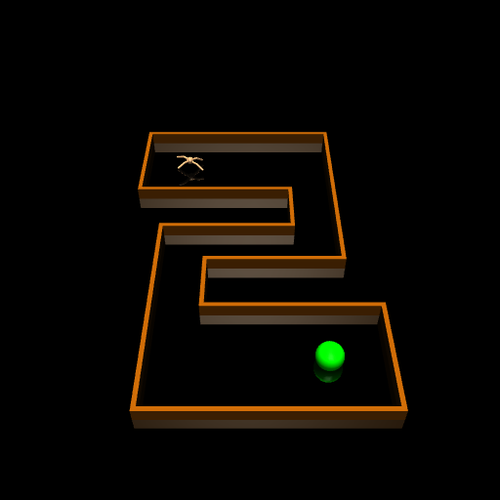}\end{subfigure}
    \begin{subfigure}[t]{0.095\textwidth}\includegraphics[width=\linewidth]{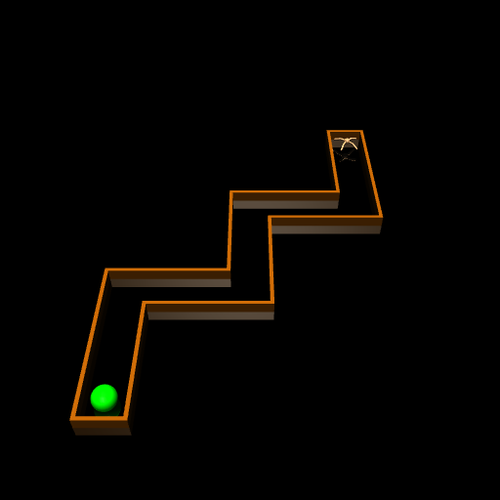}\end{subfigure}
  \end{center}
  \caption{10-Maze-v2 lifelong learning taskset}\label{10Mazev2}
\end{figure*}

\subsubsection{Oracle Gating Controller}
One might suspect that all gains in sample efficiency come from hand-crafted model primitives because they allow the agent to learn a perfect gating controller.
However, Figure~\ref{fig:oracle} shows the reward curves for an experiment where the gating controller is already perfectly known. 
This setup is unable to learn any 10-Maze task. 
Since the 10-Maze taskset is composed of sequential subtasks, only one subpolicy will be learned in the first corridor when the gating controller is perfect. 
When transitioning to the second corridor, the second policy needs to be learned from scratch, making the ant's survival rate very low.
This discourages the first subpolicy from entering the second corridor and activating the second subpolicy.
Eventually, the ant stops moving forward close to the intersection between the first two corridors.
In contrast, MPHRL's natural curriculum for gradual specialization allows multiple subpolicies to learn the basic skills for survival initially.

\subsubsection{Partial Decomposition}
To confirm that the ordering of tasks does not significantly affect MPHRL's performance, we modified 10-Maze to create the 10-Maze-v2 taskset (Figure~\ref{10Mazev2}), in which
the source task does not allow for complete decomposition into all useful subpolicies for the subsequent tasks.
Again, we observe large improvement in sample efficiency over standard PPO (Figure~\ref{fig:learnmp}).

\subsubsection{Learned Model Primitives}
\begin{figure}
  \begin{center}
    \begin{subfigure}[t]{0.22\columnwidth}\includegraphics[width=\linewidth]{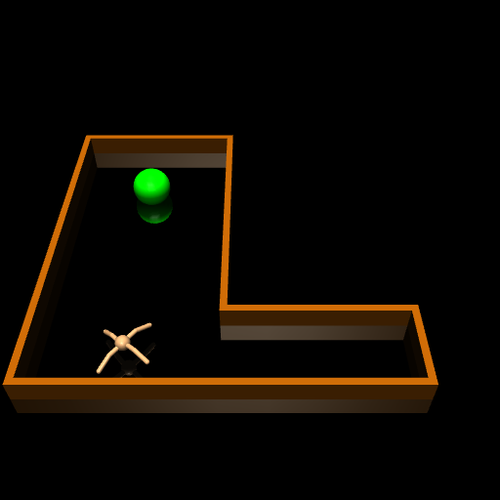}\end{subfigure}\hspace{\baselineskip}
    \begin{subfigure}[t]{0.22\columnwidth}\includegraphics[width=\linewidth]{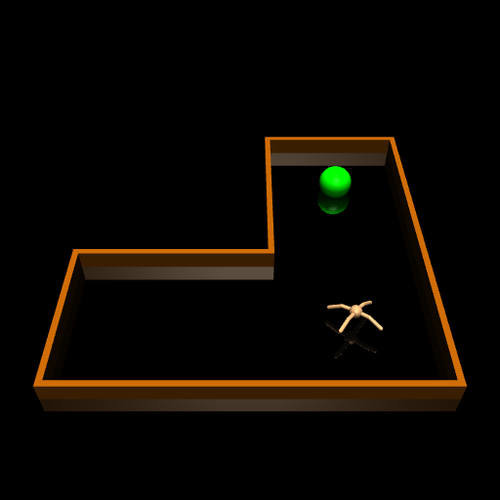}\end{subfigure}\hspace{\baselineskip}
    \begin{subfigure}[t]{0.22\columnwidth}\includegraphics[width=\linewidth]{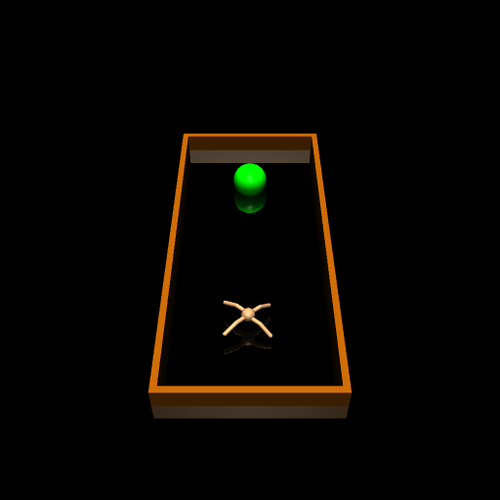}\end{subfigure}
  \end{center}
  \caption{Three corridor environments for learning the ``N'' model primitive}\label{mpex}
\end{figure}

This paper focuses on evaluating suboptimal models for task decomposition in controlled experiments using hand-designed model primitives.
Here, we show one way to obtain each model primitive for 10-Maze-v2 using three corridor environments demonstrated in Figure~\ref{mpex}.
Concretely, we parameterize each model primitive using a multivariate Gaussian distribution. We learn the mean of  this distribution via a multi-layer perceptron using a weighted mean square error in dynamics prediction as the loss.
The standard deviation is still derived from the empirical covariance $\Sigma$ as described earlier. 
Even though the diversity in these learned model primitives is much more difficult to quantify and control, their sample efficiency substantially outperforms standard PPO and slightly underperforms hand-designed model primitives with 0 and 0.5 model noises (Figure~\ref{fig:learnmp}).
\subsubsection{Gating Controller Transfer}
To explore factors that lead to negative transfer, we tested MPHRL without re-initializing the gating controller in target tasks, as shown in Figure~\ref{fig:transfer}.
Although the mean sample efficiency remains stable, its standard deviation increases dramatically, indicating volatility due to negative transfer.

\subsubsection{Subpolicy Transfer}
To measure how much gain in sample efficiency MPHRL has achieved by transferring subpolicies alone, we conducted a 10-Maze experiment by re-initializing all network weights for every new task. As shown in Figure~\ref{fig:transfer}, sample complexity more than quintuples when subpolicies are re-initialized (in \textcolor{nice-green}{green}).

\subsubsection{Coupling between Cross Entropy and Action Distribution}
To validate using $\hat{P}(M_k \mid s_t, a_t, s_{t+1})$ in Eq.~\ref{eq:phat} as opposed to $P(M_k \mid s_t, a_t, s_{t+1})$ from Eq.~\ref{eq:posterior}, we tested MPHRL with Eq.~\ref{eq:posterior} on 10-Maze.
All runs with different seeds failed to solve the first 5 tasks (Table~\ref{tbl:coupling}).
As the gating controller is re-initialized during transfer, most actions were chosen incorrectly.
The gating controller is thus presented with the incorrect cross entropy target, which worsens the action distribution.
The resulting vicious cycle forces the gating controller to converge to a suboptimal equilibrium against the incorrect target.
\begin{table}[H]
  \setlength\tabcolsep{5pt}
  \centering
  \caption{10-Maze: effect of coupling between cross entropy and action distribution}
    \resizebox{\linewidth}{!}{\begin{tabular*}{0.65\textwidth}{lSSSSSc}
      \toprule
       & \multicolumn{5}{c}{Timesteps to reach $80\%$ average success rate ($\times 10^6$)} \\ \cmidrule(l){2-6}
       Task & 1 & 2 & 3 & 4 & 5 \\ \midrule
       Timesteps ($\times 10^6$)& 15.5 \pm 1.2& 3.4 \pm 1.1& 19.3 \pm 8.0& 28.9 \pm 2.5 & N/A \\
      \bottomrule
    \end{tabular*}}
    \label{tbl:coupling}
\end{table}
\section{Conclusions}
We showed how imperfect world models can be used to decompose a complex task into simpler ones. 
We introduced a framework that uses these model primitives to learn piecewise functional decompositions of solutions to complex tasks.
The learned decomposed subpolicies can then be used to transfer to a variety of related tasks, reducing the overall sample complexity required to learn complex behaviors.
Our experiments showed that such structured decomposition avoids negative transfer and catastrophic interference, a major concern for lifelong learning systems.

Our approach does not require access to accurate world models.
Neither does it need a well-designed task distribution or the incremental introduction of individual tasks.
So long as the set of model primitives are useful across the task distribution, MPHRL is robust to other imperfections.

Nevertheless, learning useful and diverse model primitives, subpolicies and task decomposition \textit{all simultaneously} is left for future work.
The recently introduced Neural Processes~\cite{Garnelo2018-kn} can potentially be an efficient approach to build upon. 

\section*{Acknowledgments}
We are thankful to Kunal Menda and everyone at SISL for useful comments and suggestions. This work is supported in part by DARPA under agreement number D17AP00032.
The content is solely the responsibility of the authors and does not necessarily represent the official views of DARPA.
We are also grateful for the support from Google Cloud in scaling our experiments. 
\pagebreak
\bibliographystyle{ACM-Reference-Format}  %
\bibliography{references}
\end{document}